\documentclass[11pt]{article}

\usepackage[final]{acl}

\usepackage{times}
\usepackage{latexsym}

\usepackage[T1]{fontenc}
\usepackage[utf8]{inputenc}

\usepackage{microtype}

\usepackage{inconsolata}

\usepackage{graphicx}
\usepackage{stackengine}

\usepackage{amssymb}
\usepackage{multirow}
\usepackage{makecell}

\usepackage{subfigure}
\usepackage{amsthm}
\usepackage[section]{placeins} 
\usepackage{array}
\usepackage{amsmath}  
\usepackage{algpseudocode}
\usepackage[dvipsnames]{xcolor}
\usepackage[most]{tcolorbox}
\usepackage{tabularx}
\usepackage{algorithm}
\usepackage{algpseudocode}
\usepackage{tcolorbox} % For tcolorbox environment
\usepackage{listings}  % For lstlisting environment
\usepackage[table]{xcolor}
\usepackage{colortbl}
\usepackage{enumitem}
\usepackage{changepage}
\usepackage{pifont}

\usepackage{siunitx}

\hypersetup{
  colorlinks,
  citecolor=blue,
  linkcolor=blue,
  urlcolor=blue
}

\definecolor{cellgray}{gray}{0.92}

\newcommand{\varcell}[2]{#1 {\tiny $\pm$ #2}}

\title{Language-Grounded Multi-Agent Planning for Personalized and Fair Participatory Urban Sensing}

\author{
 \textbf{Xusen Guo\textsuperscript{1}},
 \textbf{Mingxing Peng\textsuperscript{1}},
 \textbf{Hongliang Lu\textsuperscript{2}},
 \textbf{Hai Yang\textsuperscript{1, 2}},
\\
 \textbf{Jun Ma\textsuperscript{1}},
 \textbf{Yuxuan Liang\textsuperscript{1*}},
\\
\\
 \textsuperscript{1}The Hong Kong University of Science and Technology (Guangzhou) \\
 \textsuperscript{2}The Hong Kong University of Science and Technology
\\
 \small{
   \textbf{Correspondence:} \href{yuxliang@outlook.com}{yuxliang@outlook.com}
 }
}

\begin{document}
\maketitle

\begin{abstract}
Participatory urban sensing leverages human mobility for large-scale urban data collection, yet existing methods typically rely on centralized optimization and assume homogeneous participants, resulting in rigid assignments that overlook personal preferences and heterogeneous urban contexts. We propose \textit{MAPUS}, an LLM-based multi-agent framework for personalized and fair participatory urban sensing. In our framework, participants are modeled as autonomous agents with individual profiles and schedules, while a coordinator agent performs fairness-aware selection and refines sensing routes through language-based negotiation. Experiments on real-world datasets show that \textit{MAPUS} achieves competitive sensing coverage while substantially improving participant satisfaction and fairness, promoting more human-centric and sustainable urban sensing systems.
\end{abstract}

\section{Introduction}

\begin{figure*}
    \centering
    \includegraphics[trim={2cm 10cm 3cm 9cm}, clip, width=0.84\linewidth]{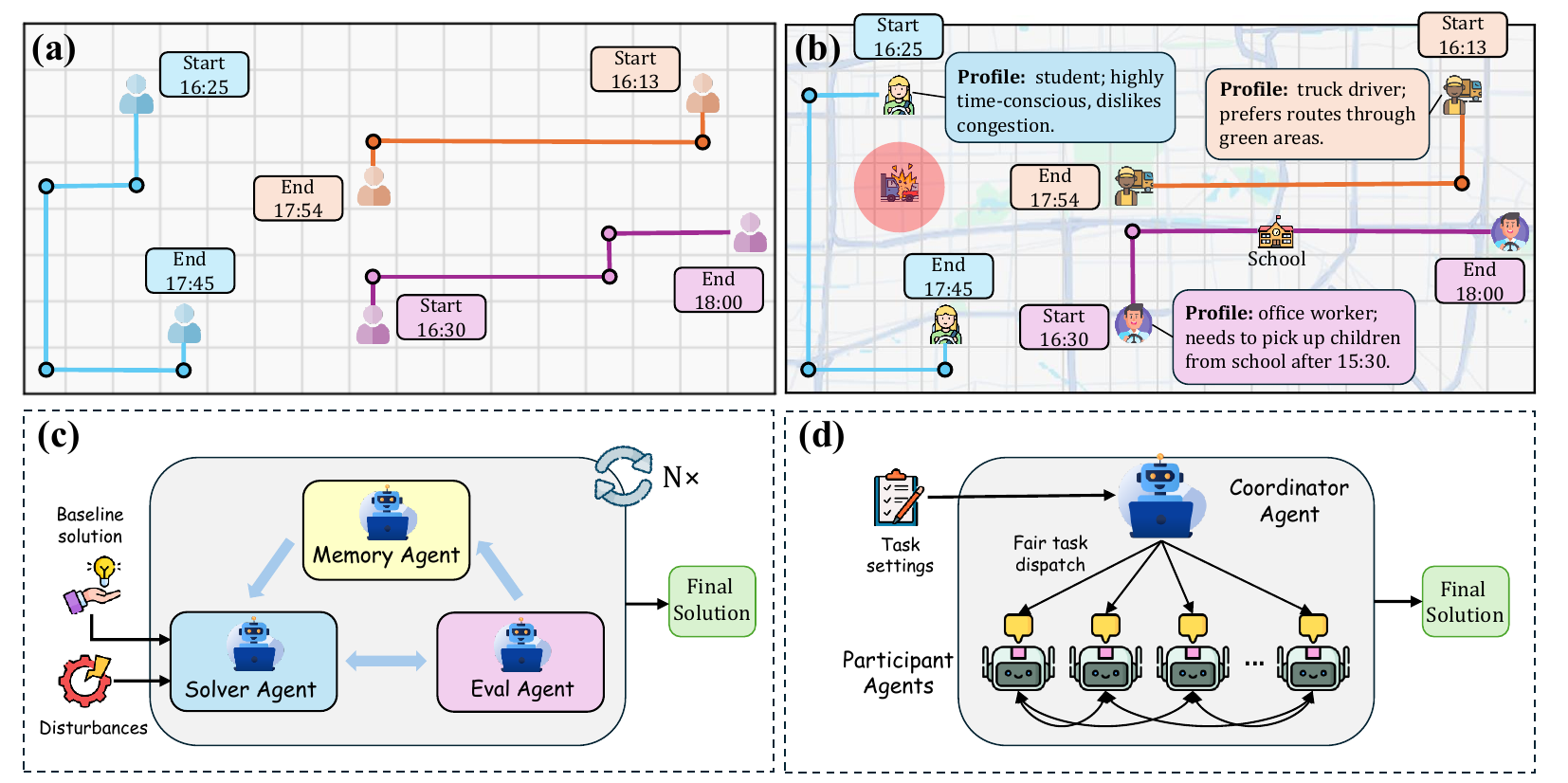}
    \caption{Comparison of paradigms for PUS problem. \textbf{(a)} Conventional centralized PUS relies on simplified assumptions about participants and sensing regions. \textbf{(b)} MAPUS models participants as autonomous agents with personalized preferences while incorporating heterogeneous urban context. \textbf{(c)} AgentSense \citep{guo2025agentsense} uses LLM agents as workflow components to iteratively refine a baseline PUS solution under disturbances. \textbf{(d)} MAPUS adopts a coordinator-and-participants multi-agent paradigm, enabling decentralized, personalized, and fairness-aware planning.}
\label{fig:figure1}
\vspace{-1em}
\end{figure*}

Understanding urban dynamics increasingly relies on large-scale sensing data collected across cities. Participatory urban sensing (PUS) has emerged as a promising approach to meet this need \citep{ganti2011mobile, kanhere2013participatory}. In PUS, urban commuters, such as ride-hailing drivers and daily travelers, are recruited to collect data with monetary incentives. To ensure efficient data collection, PUS requires designing sensible routes based on participants' travel itineraries. Compared to static sensor network infrastructure, PUS offers greater flexibility and much lower deployment costs, making it particularly well-suited for applications such as traffic monitoring \citep{du2014effective} and environmental sensing \citep{jiang2016citizen}.

Despite its potential, the practical deployment of PUS remains challenging. Most existing PUS methods \citep{ji2023survey, zhang2014crowdrecruiter} model this task as a combinatorial optimization problem based on data coverage, and typically assume that, aside from itinerary differences, participants are homogeneous and that sensing tasks are conducted on uniform space\citep{ding2021crowdsourcing, wang2024urban}, as illustrated in Figure~\ref{fig:figure1}(a). However, this oversimplification fails to capture the complexities of real-world \textit{human preferences} and \textit{regional attributes}, which are typically expressed in textual formats and are hard to be encoded as numerical constraints in optimization. Moreover, relying solely on maximizing data coverage can result in an imbalanced task allocation, where certain individuals are repeatedly selected, leading to workload overload and reduced fairness. Given these challenges, there is a critical need for a solution that takes into account diverse human preferences, regional attributes, and fairness considerations, ensuring that the system is inclusive and sustainable in real-world deployment.

Recently, large language models (LLMs) have shown great potential in handling complex, context-dependent planning tasks due to their ability to reason over natural language \citep{ferrag2025llm, openai2023gpt4, team2023gemini, guo2025deepseek}. Some researchers have utilized LLMs to solve combinatorial optimization problems. Existing efforts mainly follow two directions: one treats LLMs as optimizers that can handle language-based constraints \citep{yang2023large, jiang2025large}, while the other uses LLMs to explore new heuristic functions for improving existing algorithms \citep{wu2025efficient, chen2025hifo, ha2025pareto}. Among these, AgentSense \citep{guo2025agentsense} was the first to introduce LLMs to the PUS problem. It uses three LLM agents (solver, eval, and memory agent) to construct a workflow that refines a baseline solution under dynamic disturbances, as illustrated in Figure~\ref{fig:figure1}(c). While it offers a fresh perspective for solving the PUS problem, it still relies on idealized assumptions and does not address the limitations of traditional approaches. As a result, it fails to provide a fair, human-centered task planning that considers the diverse needs of participants.

In this work, we propose \textbf{MAPUS}, a \textbf{M}ulti-\textbf{A}gent framework for \textbf{P}articipatory \textbf{U}rban \textbf{S}ensing. MAPUS consists of a coordinator agent and a set of participant agents, all integrated with an LLM to enhance their reasoning and decision-making capabilities, as shown in Figure~\ref{fig:figure1}(d). The coordinator agent oversees the overall task allocation process, including task broadcasting, participant selection, and coordination among participants, while the participant agents model heterogeneous human behaviors and preferences. As illustrated in Figure~\ref{fig:figure1}(b), the participant agents enable \textit{personalized} route planning by reasoning over both individualized participant profiles and semantic information of sensing regions, such as land-use types and other urban attributes. Based on these inputs, each participant agent can decide whether to accept a task and generate a sensing route that better matches its own needs and preferences. Meanwhile, the coordinator agent promotes \textit{fairness} at the system level by adjusting task allocation according to participants' historical involvement, so that sensing opportunities are distributed more evenly across the population. Through this design, MAPUS unifies participant-level personalization and system-level fairness within a human-centered multi-agent cooperation system for PUS.

In summary, our contributions are as follows:
\begin{itemize}[leftmargin=*,itemsep=0.1em,topsep=0em]
    \item \textbf{Language-grounded multi-agent formulation for PUS.}
    We reformulate PUS as a decentralized coordination problem, where participants are modeled as autonomous LLM agents with heterogeneous preferences, and sensing tasks are embedded in heterogeneous urban context.

    \item \textbf{LLM-based preference- and context-aware route planning.}
    We develop a planning mechanism in which participant agents reason jointly over individual profiles and urban attributes to generate personalized sensing routes.

    \item \textbf{Fair and efficient multi-agent coordination.}
    We introduce a coordinator-driven mechanism that performs fairness-aware selection and negotiation-based route refinement to mitigate workload imbalance and sensing redundancy.

    \item \textbf{Comprehensive evaluation on real-world mobility data.}
    Experiments on large-scale datasets show that MAPUS improves participant satisfaction and fairness while maintaining competitive sensing utility against centralized baselines.
\end{itemize}

\section{Related Work}

Participatory urban sensing leverages mobile users as distributed sensors for large-scale urban data collection, supporting applications such as traffic monitoring and environmental sensing \citep{wang2018city, hasenfratz2012participatory}. A large body of work formulates this task as an optimization problem that maximizes spatio-temporal coverage under resource constraints \citep{ji2023survey}. Representative approaches include drive-by sensing models that select vehicle trajectories to maximize city-wide coverage \citep{ji2016urban} and probabilistic recruitment strategies that account for spatial uncertainty \citep{zhang2014crowdrecruiter}. To improve scalability and adaptivity, heuristic and learning-based approaches have also been proposed, including greedy selection based on marginal coverage gains and reinforcement learning methods for dynamic task assignment \citep{wang2024urban, ding2021crowdsourcing}. Despite improving sensing efficiency, most existing frameworks assume homogeneous participants and rely on centralized optimization, overlooking individual preferences, contextual sensitivities, and long-term workload fairness \citep{blunck2013heterogeneity, chen2020fair}. These limitations motivate more human-centric sensing frameworks that explicitly account for heterogeneous participants and various urban context.

Recent advances in large language models have enabled language-based planning and multi-agent coordination through natural language reasoning \citep{wei2022chain, yao2023tree, schick2023toolformer}. Building on these capabilities, LLM-based multi-agent systems support role specialization and structured interaction for complex decision-making tasks \citep{hong2024metagpt, guo2024large, han2024llm}. Such agents have shown strong potential in negotiation, critique, and behavior simulation \citep{aher2023using, argyle2023out}, and have recently been applied to urban systems such as participatory planning, mobility modeling, and city-scale simulations \citep{zhou2024large, wang2024large, yan2024opencity}. Some studies also explore LLMs as explainable intermediaries in participatory sensing \citep{hou2025urban, guo2025agentsense}. However, most existing approaches still rely on centralized planning or static agent roles, limiting participant-level autonomy and personalization. In contrast, our work models each participant as an autonomous LLM agent with individualized profiles, and formulates PUS as decentralized coordination among heterogeneous participants, enabling personalized route planning.

\section{Methodology}
\vspace{-0.5em}

\begin{figure*}
    \centering
    \includegraphics[trim={0cm 7cm 0cm 6cm}, clip, width=\linewidth]{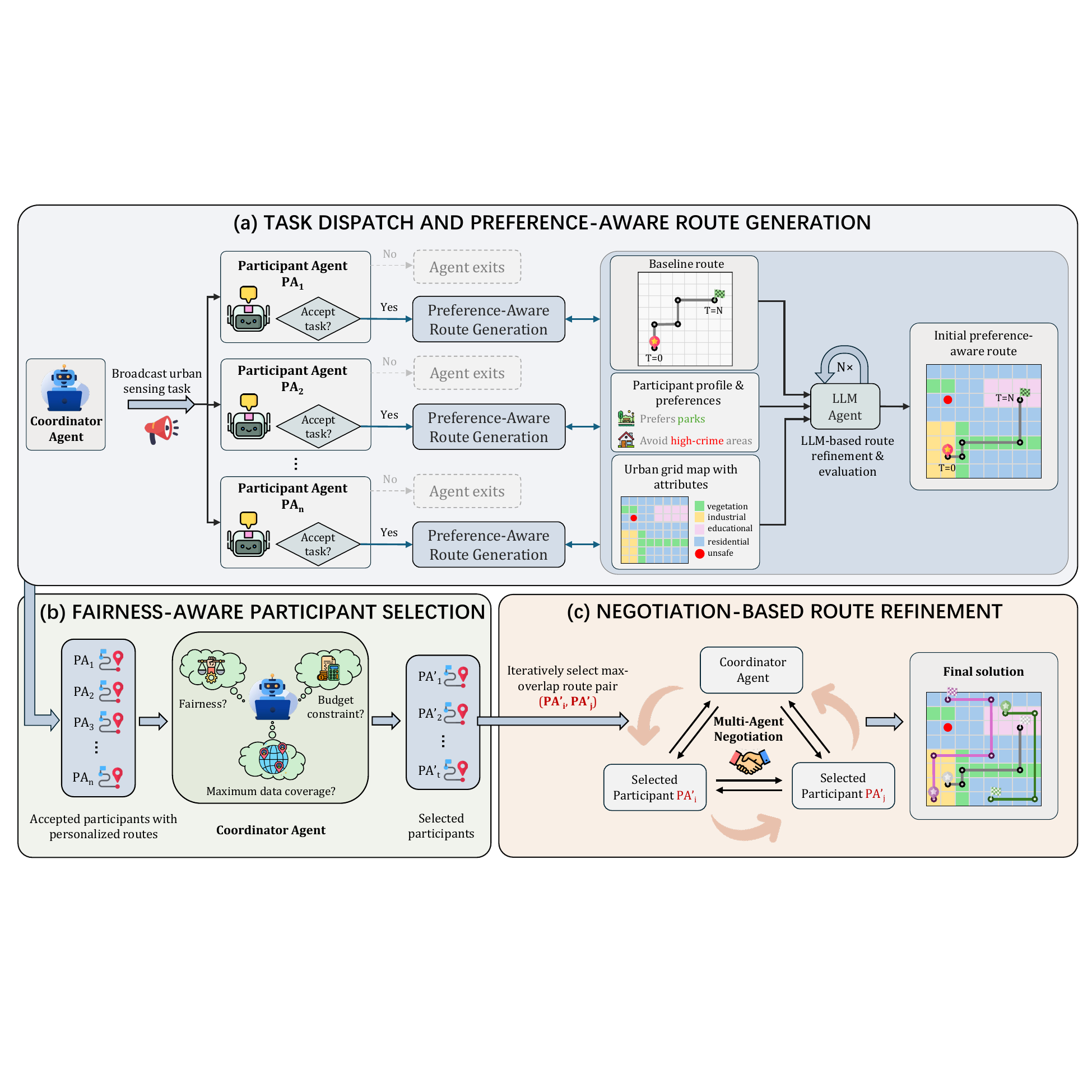}
    \caption{Overview of proposed MAPUS. It operates in three stages: (a) task dispatch and preference-aware route generation, (b) fairness-aware participant selection, and (c) negotiation-based route refinement.}
\label{fig:framework}
\vspace{-1.0em}
\end{figure*}

\subsection{Problem Definition}

We consider a participatory urban sensing task over a grid-based region and a finite time horizon. Given a set of candidate participants, the platform needs to decide which participants to recruit and what sensing route to assign to each selected participant. Each participant is associated with mobility constraints (origin, destination, departure time, arrival time, and speed), a recruitment cost, and personalized profiles, and any assigned route must satisfy the participant's mobility constraints. Let $x_u \in \{0,1\}$ indicate whether participant $u$ is recruited, and let $r_u$ denote the route assigned to participant $u$. The problem is to maximize system-level sensing utility together with participant-level route satisfaction:
\begin{equation}
\setlength{\abovedisplayskip}{6pt} % Adjust space above the equation
\setlength{\belowdisplayskip}{6pt} % Adjust space below the equation
\begin{aligned}
\max_{\{x_u, r_u\}} \quad 
& \phi(\{x_u, r_u\}) + \eta \sum_{u} x_u\, S_u(r_u) \\
\text{s.t.} \quad
& \sum_u c_u x_u \le B,
\end{aligned}
\label{eq:pus_problem}
\end{equation}
where $\phi$ denotes the utility of the collected sensing data, which is typically evaluated by its spatio-temporal coverage quality. $S_u(r_u)$ measures how well route $r_u$ matches participant $u$'s preferences, $c_u$ is the recruitment cost of participant $u$, and $B$ is the total budget. Full definitions are provided in Appendix~\ref{appendix:problem_setup}.

% \vspace{-0.7em}
\subsection{Framework Overview}

As illustrated in Figure~\ref{fig:framework}, MAPUS operates in three stages. First, the coordinator agent (CA) broadcasts the sensing task to all participant agents (PAs). Each PA then decides whether to accept the task according to the task setting and its own itinerary. For those that accept, the corresponding PAs generate preference-aware routes based on their personalized profiles and the urban attributes of the sensing regions, leveraging the reasoning capability of LLMs. Second, the CA collects the routes generated by all accepted PAs and selects a subset of participants under the budget constraint. This selection considers both the historical participation records of PAs and the overall data coverage of the final sensing plan, so as to improve fairness while maintaining high sensing utility. Finally, to further improve the coverage quality of the collected data, the CA repeatedly identifies the pair of selected PAs with the highest route overlap and brings them into a negotiation process. Through incentive-guided interaction, the PAs iteratively adjust their routes to reduce redundancy and improve the overall sensing efficiency. The following sections will present the technical details for each stage.

\subsection{Preference-Aware Route Generation}
For accepted PAs, MAPUS generates personalized sensing routes through iterative refinement. The main challenge is that PUS route planning must satisfy hard mobility constraints while accounting for soft, participant-specific preferences over urban regions, which are often semantic and difficult to encode as fixed rules or explicit objectives. MAPUS therefore combines a classical route planner with an LLM: as illustrated in Figure~\ref{fig:framework}(a), a feasible baseline route is first constructed and then iteratively refined by the LLM, which jointly jointly considers the task description, participant profile, and urban attributes. Candidate routes that violate feasibility constraints are discarded, while valid routes are accepted only if they improve the sensing utility defined in Equation~\ref{eq:pus_problem}. In this design, the classical planner ensures feasibility, while the LLM enables personalized route adaptation. See Appendix~\ref{appendix:route_generation_details} for details and the prompt design.

To evaluate whether a refined route better matches participant needs, we introduce the \emph{path satisfaction score} (PSS) to measure how well a sensing route aligns with a participant's preferences while accounting for undesirable urban conditions:
\begin{equation}
\setlength{\abovedisplayskip}{6pt} % Adjust space above the equation
\setlength{\belowdisplayskip}{6pt} % Adjust space below the equation
S(r_u) = \mathrm{sim}\big(\mathbf{p}_u,\mathbf{h}(r_u)\big) - \mu\,\mathrm{risk}(r_u),
\label{eq:path_satisfaction}
\end{equation}
The similarity function $\mathrm{sim}(\cdot,\cdot)$ measures the alignment between participant preferences $\mathbf{p}_u$ and route attributes $\mathbf{h}(r_u)$, implemented as the cosine similarity between the two normalized distributions. The term $\mathrm{risk}(r_u)$ aggregates undesirable urban exposure along the route, such as high crime regions.

Specifically, $\mathbf{h}(r_u)$ is defined as the empirical land-use distribution along the route $r_u$
\begin{equation}
\setlength{\abovedisplayskip}{5pt} % Adjust space above the equation
\setlength{\belowdisplayskip}{5pt} % Adjust space below the equation
\mathbf{h}(r_u)(k)
=
\frac{1}{L_u}\sum_{i=1}^{L_u}
\mathbb{I}\big[\kappa(g_i)=k\big],
\quad k\in\mathcal{K},
\label{eq:route_histogram}
\end{equation}
where $L_u$ is the route length and
\begin{equation}
\setlength{\abovedisplayskip}{6pt} % Adjust space above the equation
\setlength{\belowdisplayskip}{6pt} % Adjust space below the equation
\small
\begin{aligned}
\kappa(g) \in \mathcal{K} =
\{
&\texttt{vegetation},\texttt{industrial},\texttt{institutional}, \\
&\texttt{medical},\texttt{residential},\texttt{commercial}\},
\end{aligned}
\label{eq:landuse_category}
\end{equation}
denotes six land-use categories of grid cell $g$, derived from points of interest (POIs) within the grid. $\mathbf{p}_u$ is represented as a normalized distribution over $\mathcal{K}$, where each element $\mathbf{p}_u(k)$ reflects participant $u$'s relative preference for traversing regions of type $k$. This representation provides a compact and interpretable abstraction of individual route preferences and can be estimated from historical mobility traces or stated preference information. 

\subsection{Fairness-Aware Participant Selection}

After candidate PAs submit their feasible routes, the CA incrementally selects a subset of them under a limited recruitment budget, as illustrated in Figure~\ref{fig:framework}(b). At each step, only PAs whose recruitment cost can be covered by the remaining budget are considered. Instead of selecting participants solely based on data coverage gain, MAPUS balances sensing utility with long-term participation fairness. Specifically, for each budget-feasible participant $u$, the coordinator first computes the marginal data coverage gain
\begin{equation}
\setlength{\abovedisplayskip}{5pt} % Adjust space above the equation
\setlength{\belowdisplayskip}{5pt} % Adjust space below the equation
\Delta \phi(u) = \phi\!\left(\bigcup_{w\in \mathcal{S}\cup\{u\}} r_w\right) - \phi\!\left(\bigcup_{w\in \mathcal{S}} r_w\right),
\label{eq:marginal_gain}
\end{equation}
where $\mathcal{S}$ denotes the current set of selected PAs. To promote long-term fairness across tasks, the coordinator also assigns a fairness score $F(u)=\frac{1}{1+h_u}$, where $h_u$ is the historical number of times participant $u$ has been selected. After normalization, the two components are combined into a composite selection score
\begin{equation}
\setlength{\abovedisplayskip}{5pt} % Adjust space above the equation
\setlength{\belowdisplayskip}{5pt} % Adjust space below the equation
J(u)=\beta\cdot \widetilde{\Delta \phi}(u) + (1-\beta)\cdot \widetilde{F}(u),
\label{eq:selection_score}
\end{equation}
where $\beta\in[0,1]$ controls the trade-off between coverage efficiency and fairness.

CA greedily selects participants according to $J(u)$ and updates the remaining budget after each selection until no additional participant can be recruited. When multiple candidate PAs obtain nearly identical scores, MAPUS invokes an LLM-based tie-breaking mechanism. Compared with a rule-based scheme, the LLM can more flexibly reason over richer information inferred from participant profiles, such as task suitability and economic situation, when making the final selection. This enables fairer participant selection beyond purely score-based ranking. Additional details and LLM prompts are provided in Appendix~\ref{appendix:selection_details}.

\vspace{-0.5em}
\subsection{Negotiation-Based Route Refinement}

Although the participant selection stage balances sensing coverage and fairness, independently generated routes may still exhibit substantial spatial overlap, leading to redundant sensing and inefficient use of participant resources. To address this issue, MAPUS introduces a negotiation-based route refinement process, in which the CA identifies highly overlapping route pairs and mediates route adjustments among selected PAs. Given the selected routes $\{r_u \mid u \in \mathcal{S}\}$, the CA first computes pairwise route overlap and then iteratively selects the most overlapping pair for refinement. The overlap between routes $r_u$ and $r_v$ is defined as
\begin{equation}
\setlength{\abovedisplayskip}{5pt} % Adjust space above the equation
\setlength{\belowdisplayskip}{5pt} % Adjust space below the equation
\small
\mathrm{Overlap}(r_u,r_v)=
\frac{
\sum_{g\in\mathcal{G}} \cdot \min\big(\mathbf{1}[g\in r_u],\mathbf{1}[g\in r_v]\big)
}{
\sum_{g\in\mathcal{G}} \cdot \max\big(\mathbf{1}[g\in r_u],\mathbf{1}[g\in r_v]\big)
}.
\label{eq:route_overlap_main}
\end{equation}
This formula measures the relative redundancy between two routes and remains comparable across different lengths. At each iteration, the CA selects the route pair with the highest overlap and proposes revisions that reduce route overlap while preserving route feasibility.

It is worth noting that this negotiation process is consent-driven. In each round, the CA proposes refined routes to reduce route overlap, and PAs evaluate the proposal according to their own preferences and constraints. If both PAs accept, the refined routes are adopted; otherwise, the CA updates the proposal based on their feedback and continues the negotiation. If no agreement is reached, the CA moves to the next most overlapping pair. Through this iterative process, MAPUS reduces sensing redundancy while preserving decentralized decision-making and participant autonomy. Additional details and the negotiation prompt are provided in Appendix~\ref{appendix:negotiation_details}.

\section{Experiments and Results}
\vspace{-0.5em}
\subsection{Experimental Setup}

\paragraph{Datasets.}
We evaluate the proposed framework on two large-scale real-world mobility datasets, T-Drive~\citep{tdriver2010} and Grab-Posisi~\citep{grabposis2019}, which represent distinct urban environments and sensing densities. 
The T-Drive dataset contains GPS trajectories from more than 10,000 taxis in Beijing collected over one week. The trajectories cover a large metropolitan region with heterogeneous land-use patterns and relatively sparse sampling intervals. The Grab-Posisi dataset contains high-frequency positional data from ride-hailing drivers across Southeast Asian cities. The dataset was collected in April 2019 with a sampling rate of approximately 1 second and includes dense trajectories enriched with contextual attributes. Compared with T-Drive, Grab-Posisi features much denser mobility traces and more localized movement patterns. 

\paragraph{Urban Attributes and Participant Profiles.}
To capture region-level contextual signals in urban sensing tasks, we incorporate land-use distributions and crime statistics as grid-level attributes. The former is inferred from POIs, whereas the crime statistics are obtained from publicly available datasets~\citep{zhang2025llm}. To model heterogeneous participant behaviors, we further construct personalized participant profiles that include economic status, behavioral traits, and land-use preferences. These profiles are used to instantiate participant agents and guide their task acceptance decisions and route planning. Details of these components are provided in Appendix~\ref{appendix:urban_attributes} and~\ref{appendix:participant_profiles}.

\paragraph{Task Settings.}

\begin{table}[!t]
\scriptsize
\centering
\renewcommand{\arraystretch}{1.1}
\begin{tabular}{p{1.0cm}|p{0.65cm}|c|c|c|c}
    \Xhline{1.0pt} 
    Dataset & Scale & \# Regions & $N$ & $B$ & $H$ (min)\\
    \Xhline{1.0pt} 
    \multirow{3}{*}{T-Drive} 
    & Small  & 64 (8 $\times 8$) & 20 & 40  & 120\\
    & Medium & 256 (16 $\times 16$) & 40 & 60  & 240\\
    & Large  & 1024 (32 $\times 32$) & 60 & 100 & 360\\
    \hline
    \multirow{3}{*}{Grab-Posisi} 
    & Small  & 32 (8 $\times$ 4) & 15 & 40  & 40\\
    & Medium & 128 (16 $\times 8$) & 30 & 60  & 80\\
    & Large  & 512 (32 $\times 16$) & 45 & 100 & 160\\
    \Xhline{1.0pt} 
\end{tabular}
\caption{Experimental configurations under different scales for the two datasets. $\#\text{Regions}$ denotes the number of spatial sensing regions, $N$ the number of candidate participants, $B$ the total budget, and $H$ the temporal span horizon of the sensing task.}
    \vspace{-1.0em}
\label{tab:dataset_settings}
\vspace{-1.0em}
\end{table}

To evaluate MAPUS under different scales, we construct three configurations for each dataset, namely \emph{Small}, \emph{Medium}, and \emph{Large}, as summarized in Table~\ref{tab:dataset_settings}. 
The urban area is discretized into a spatial grid and sensing tasks are defined over a finite time horizon, forming spatio-temporal sensing regions. Following prior work~\citep{ji2016urban, wang2024urban}, participants move at constant speed and each sensing action yields a normalized reward of 1.0 per time step. The sensing interval is set to 15 minutes for T-Drive and 5 minutes for Grab-Posisi to match their sampling granularities. For each configuration, we generate 20 independent task instances, and results are averaged across runs. Unless otherwise stated, both the CA and PAs in MAPUS are powered by \texttt{gpt-4.1-mini-2025-04-14}. We implement them using the LangGraph framework, which facilitates tool calling and agent memory management.

\subsection{Main Results}

Tables~\ref{tab:tdrive_overall} and~\ref{tab:grab_overall} report the overall performance on the T-Drive and Grab-Posisi datasets under three different task scales. We compare MAPUS with six representative baseline planners (details in Appendix~\ref{appendix:baselines}) and an additional single-LLM  baseline based on \texttt{gpt-4.1-mini-2025-04-14}, using two metrics: data coverage utility and path satisfaction score (PSS). In this single-LLM  setting, all information about task settings, mobility constraints, participant profiles and regional attributes are integrated and provided to one LLM, which directly generates the final sensing plan without the explicit task decomposition and coordination.

\paragraph{Data Coverage utility.} 
The results show that MAPUS achieves competitive or superior coverage across both datasets. In the \emph{Small} setting, most methods obtain similar coverage, since the limited problem size allows classical planners to explore the solution space effectively. For example, on T-Drive (\emph{Small}), GraphDP achieves slightly higher coverage (4.761) than MAPUS (4.677). However, as task scale increases, MAPUS consistently achieves the best or near-best coverage. On T-Drive, MAPUS obtains the highest coverage at both \emph{Medium} (5.115) and \emph{Large} (5.251) scales. On Grab-Posisi, coverage differences are modest for smaller scales, while MAPUS achieves the highest coverage in the \emph{Large} setting (5.701). MAPUS also consistently outperforms the single-LLM  baseline, suggesting that multi-agent decomposition is more effective when the model must handle large amounts of participant and regional information.

\begin{table*}[!t]
\centering
\small
\renewcommand{\arraystretch}{1.0}
\begin{tabular}{c|cc|cc|cc}
\Xhline{1.2pt}
\multirow{3}{*}{Method} 
& \multicolumn{6}{c}{Task Scales} \\
\cline{2-7}
& \multicolumn{2}{c|}{Small}
& \multicolumn{2}{c|}{Medium}
& \multicolumn{2}{c}{Large} \\
\cline{2-7}
& Coverage & PSS
& Coverage & PSS
& Coverage & PSS \\
\Xhline{1.2pt}
RN     
& \varcell{4.557}{8.38e-3} & \varcell{0.632}{9.60e-5}
& \varcell{4.626}{1.46e-2} & \varcell{0.605}{4.65e-5}
& \varcell{4.974}{3.68e-2} & \varcell{0.594}{3.10e-4}\\

TVPG   
& \varcell{4.563}{1.34e-2} & \varcell{0.634}{9.68e-5}
& \varcell{4.753}{2.27e-2} & \varcell{0.610}{6.62e-5}
& \varcell{5.067}{1.34e-2} & \varcell{0.590}{2.14e-4} \\

TCPG   
& \varcell{4.594}{7.31e-3} & \varcell{0.631}{4.58e-5}
& \varcell{5.084}{1.13e-2} & \varcell{0.621}{3.48e-5}
& \varcell{5.043}{1.71e-2} & \varcell{0.602}{1.17e-4}\\

MSA    
& \varcell{4.662}{5.60e-3} & \varcell{0.630}{3.60e-5}
& \varcell{5.065}{9.24e-3} & \varcell{0.622}{8.63e-5}
& \varcell{5.131}{1.70e-2} & \varcell{0.598}{1.80e-4}\\

MSAGI  
& \varcell{4.661}{5.63e-3} & \varcell{0.633}{6.57e-5}
& \varcell{5.052}{9.98e-3} & \varcell{0.616}{7.08e-5}
& \varcell{5.128}{1.54e-2} & \varcell{0.590}{6.43e-5}\\

GraphDP
& \varcell{\textbf{4.761}}{2.68e-3} & \varcell{0.659}{1.98e-4}
& \varcell{5.102}{1.86e-3} & \varcell{0.651}{2.93e-4}
& \varcell{5.214}{2.13e-2} & \varcell{0.628}{5.05e-4}\\

GPT-4.1-mini
& \varcell{4.576}{\textcolor{red}{0.67}} & \varcell{0.742}{\textcolor{red}{0.054}}
& \varcell{4.872}{\textcolor{red}{0.58}} & \varcell{0.706}{\textcolor{red}{0.084}}
& \varcell{4.936}{\textcolor{red}{0.72}} & \varcell{0.691}{\textcolor{red}{0.045}} \\

\textbf{MAPUS (Ours)}
& \varcell{4.677}{0.16} & \varcell{\textbf{0.875}}{0.023}
& \varcell{\textbf{5.115}}{0.20} & \varcell{\textbf{0.875}}{0.033}
& \varcell{\textbf{5.251}}{0.18} & \varcell{\textbf{0.898}}{0.019} \\
\Xhline{1.2pt}
\end{tabular}
\caption{Overall performance comparison on the \textbf{T-Drive} dataset (20 task instances per scale). Values denote averages, with variances shown in small numbers. The largest variance per column is highlighted in red.}
\label{tab:tdrive_overall}
\vspace{-1em}
\end{table*}

\begin{table*}[!t]
\centering
\small
\renewcommand{\arraystretch}{1.0}
\begin{tabular}{c|cc|cc|cc}
\Xhline{1.2pt}
\multirow{3}{*}{Method} 
& \multicolumn{6}{c}{Task Scales} \\
\cline{2-7}
& \multicolumn{2}{c|}{Small}
& \multicolumn{2}{c|}{Medium}
& \multicolumn{2}{c}{Large} \\
\cline{2-7}
& Coverage & PSS
& Coverage & PSS
& Coverage & PSS\\
\Xhline{1.2pt}
RN     
& \varcell{4.236}{4.23e-3} & \varcell{0.727}{4.29e-5}
& \varcell{4.710}{1.18e-2} & \varcell{0.696}{6.30e-5}
& \varcell{5.399}{1.11e-2} & \varcell{0.648}{5.13e-5} \\

TVPG   
& \varcell{4.342}{4.10e-3} & \varcell{0.714}{4.59e-5}
& \varcell{4.737}{1.32e-2} & \varcell{0.695}{2.04e-4}
& \varcell{5.406}{2.62e-2} & \varcell{0.640}{5.74e-5}\\

TCPG   
& \varcell{4.232}{2.83e-3} & \varcell{0.730}{2.86e-5}
& \varcell{4.806}{1.14e-2} & \varcell{0.703}{6.40e-5}
& \varcell{5.587}{8.39e-3} & \varcell{0.658}{3.56e-5}\\

MSA    
& \varcell{4.372}{2.56e-3} & \varcell{0.744}{3.54e-5}
& \varcell{4.900}{4.15e-3} & \varcell{0.699}{9.28e-5}
& \varcell{5.586}{8.20e-3} & \varcell{0.655}{4.07e-5} \\

MSAGI  
& \varcell{\textbf{4.392}}{2.92e-3} & \varcell{0.752}{5.16e-5}
& \varcell{4.887}{4.98e-3} & \varcell{0.714}{8.49e-5}
& \varcell{5.567}{8.84e-3} & \varcell{0.653}{4.63e-5}\\

GraphDP
& \varcell{4.369}{1.69e-3} & \varcell{0.749}{4.41e-5}
& \varcell{\textbf{4.913}}{3.19e-3} & \varcell{0.719}{1.85e-4}
& \varcell{5.611}{6.41e-3} & \varcell{0.700}{1.16e-4} \\

GPT-4.1-mini
& \varcell{4.293}{\textcolor{red}{0.44}} & \varcell{0.782}{\textcolor{red}{0.032}}
& \varcell{4.755}{\textcolor{red}{0.51}} & \varcell{0.742}{\textcolor{red}{0.028}}
& \varcell{5.413}{\textcolor{red}{0.55}} & \varcell{0.729}{\textcolor{red}{0.022}} \\

\textbf{MAPUS (Ours)}
& \varcell{4.386}{0.19} & \varcell{\textbf{0.854}}{0.0087}
& \varcell{4.890}{0.26} & \varcell{\textbf{0.834}}{0.010}
& \varcell{\textbf{5.701}}{0.30} & \varcell{\textbf{0.831}}{0.012} \\
\Xhline{1.2pt}
\end{tabular}
\caption{Overall performance comparison on the \textbf{Grab-Posisi} dataset (20 task instances per scale). Values denote averages, with variances shown in small numbers. The largest variance per column is highlighted in red.}
\label{tab:grab_overall}
\vspace{-1em}
\end{table*}

\paragraph{Path satisfaction score (PSS).} 
MAPUS consistently achieves the highest PSS across all datasets and task scales, demonstrating its ability to generate routes that better align with participant preferences. On T-Drive, MAPUS improves PSS by roughly 20\%--30\% compared to the strongest baseline. The improvement on Grab-Posisi is smaller but still consistent. This difference arises from the available urban attributes used in PSS computation. In T-Drive, PSS incorporates both land-use information and crime distribution, providing richer semantic signals that our LLM-based planner can exploit. In contrast, Grab-Posisi only includes land-use attributes, which reduces the diversity of preference signals. Importantly, MAPUS maintains stable PSS across all scales, indicating that the preference-aware planning mechanism remains effective even as routing complexity increases. We can see that MAPUS also achieves much higher PSS than the single-LLM  baseline, showing that participant-level agent reasoning is more effective for personalization than directly generating a full sensing plan with one model.

\paragraph{Variance and stability.}
The variance statistics in Tables~\ref{tab:tdrive_overall} and~\ref{tab:grab_overall} provide additional insight into result stability. Compared with traditional optimization-based baselines, MAPUS generally exhibits larger variance, which is expected given the stochasticity introduced by LLM-based multi-agent reasoning and negotiation. However, MAPUS is still substantially more stable than the single-LLM  baseline, whose variance is consistently the largest. This suggests that although multi-agent LLM coordination introduces more variability than deterministic planners, the structured decomposition in MAPUS, including coordinator-guided selection and negotiation, significantly improves the stability.

\subsection{Ablation Studies}

\begin{figure}[!t]
    \centering
    \vspace{-0.8em}
    \includegraphics[trim={0cm 0cm 0cm 0cm}, clip, width=0.95\linewidth]{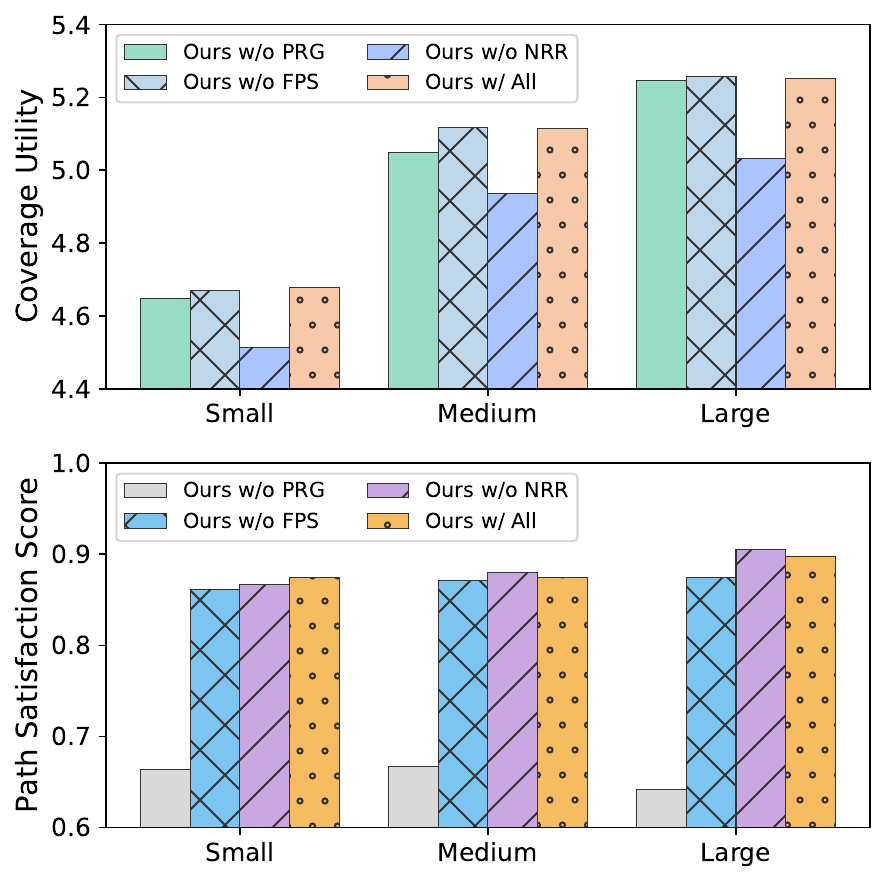}
     \caption{Ablation study on the T-Drive dataset. We compare four variants of MAPUS across three different settings: w/o PRG, w/o FPS, w/o NRR, and the full model. Top: coverage utility. Bottom: path satisfaction score (PSS).}
\label{fig:ablation}
\vspace{-1.2em}
\end{figure}

We conduct ablation experiments on the T-Drive dataset to evaluate the contribution of the three main components in MAPUS: \textit{Preference-aware Route Generation} (PRG), \textit{Fairness-aware Participant Selection} (FPS), and \textit{Negotiation-based Route Refinement} (NRR). Figure~\ref{fig:ablation} reports the coverage utility and path satisfaction score (PSS) under \emph{Small}, \emph{Medium}, and \emph{Large} settings. Removing PRG (w/o PRG), where routes are generated using the TVPG baseline, results in a slight decrease in coverage but a substantial drop in PSS across all scales. This indicates that optimization-based planners can still achieve competitive coverage, but fail to account for participant preferences. In contrast, PRG explicitly integrates preference signals during route generation, leading to significantly higher route satisfaction. Removing FPS (w/o FPS) adopts a purely greedy participant selection strategy that maximizes coverage. While this slightly improves coverage utility, it reduces PSS, as the same high-coverage participants are repeatedly selected, leading to less balanced assignments and lower participant satisfaction. Removing NRR (w/o NRR) leads to the most noticeable decrease in coverage. Without negotiation-based coordination, routes are optimized independently, resulting in substantial overlap and reduced effective sensing coverage. Meanwhile, PSS increases slightly because participants no longer need to adjust their routes to avoid conflicts.

To further analyze NRR, Table~\ref{tab:negotiation_effect} compares route statistics before and after negotiation under the \emph{Medium} setting on both datasets. The results illustrate that NRR significantly reduces route overlap (from 3.482\% to 1.267\% on T-Drive and from 5.418\% to 3.843\% on Grab-Posisi), confirming its effectiveness in mitigating redundant sensing. Although the total covered count remains similar, spatial entropy increases, suggesting that the improvement in coverage utility arises from a more efficient redistribution of sensing effort rather than from covering additional regions. A small decrease in PSS is also observed, reflecting the trade-off between global coordination and individual route preference. Overall, these results demonstrate that NRR improves global sensing efficiency while only slightly affecting individual satisfaction.

\begin{table}[!t]
\centering
\footnotesize
\renewcommand{\arraystretch}{1.0}
\begin{tabular}{l|cc|cc}
\Xhline{1.2pt}
\multirow{3}{*}{Metric}
& \multicolumn{2}{c|}{T-Drive}
& \multicolumn{2}{c}{Grab-Posisi} \\
\cline{2-5}
& \makecell{Before \\ Nego.} & \makecell{After \\ Nego.}
& \makecell{Before \\ Nego.} & \makecell{After \\ Nego.} \\
\Xhline{1.2pt}
Avg. Overlap (\%)
& 3.482 & \cellcolor{gray!15}1.267
& 5.418 & \cellcolor{gray!15}3.843 \\

Avg. Coverage
& 4.977 & \cellcolor{gray!15}5.115
& 4.770 & \cellcolor{gray!15}4.890 \\

Avg. Entropy
& 2.377 & \cellcolor{gray!15}2.477
& 2.498 & \cellcolor{gray!15}2.584 \\

Avg. Count
& 149.3 & \cellcolor{gray!15}149.3
& 132.1 & \cellcolor{gray!15}132.1 \\

Avg. PSS
& 0.880 & \cellcolor{gray!15}0.875
& 0.838 & \cellcolor{gray!15}0.834 \\
\Xhline{1.2pt}
\end{tabular}
\caption{Effect of negotiation-based route refinement under the \emph{Medium} setting (20 tasks). We report overlap, coverage utility, entropy, covered count, and PSS before and after negotiation on T-Drive and Grab-Posisi.}
\label{tab:negotiation_effect}
\vspace{-1.5em}
\end{table}

\subsection{Fairness analysis}
\label{sec5_4:fairness}

\begin{figure}
    \centering
    \vspace{-0.5em}
    \includegraphics[width=\linewidth]{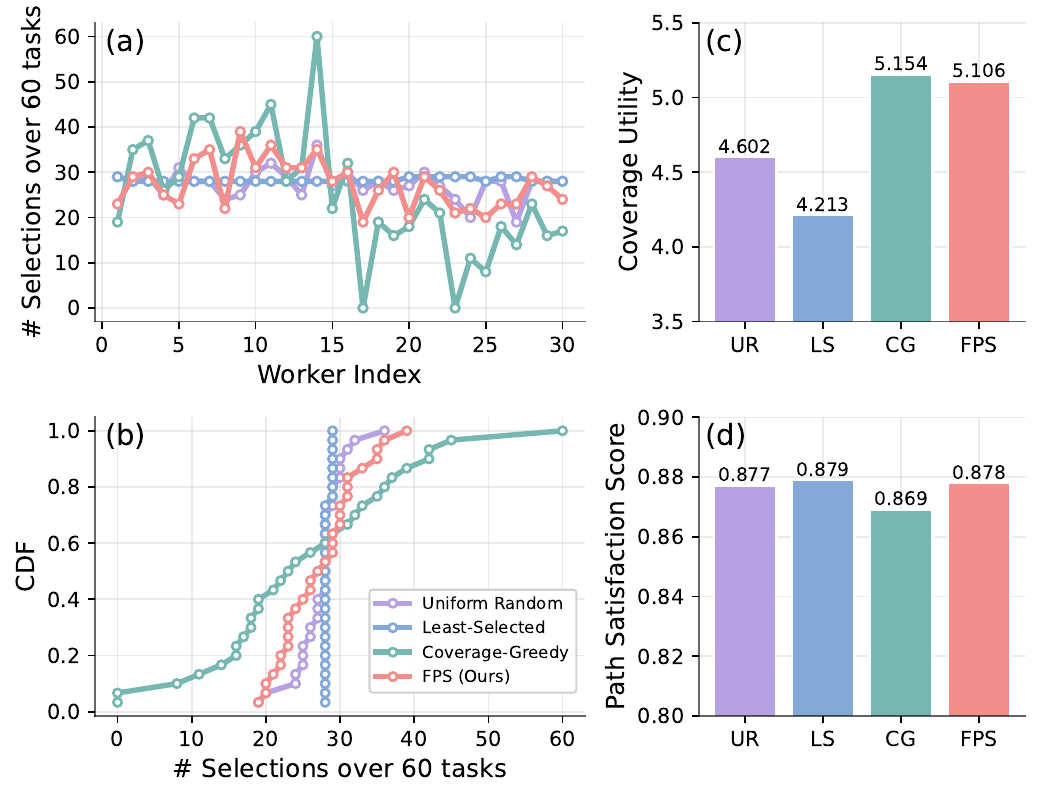}
     \caption{Fairness-aware participant selection analysis on the T-Drive dataset under the \emph{Medium} setting. (a) Per-worker selection counts over 60 tasks, (b) CDF of selection counts,  (c) data coverage utility, and (d) path satisfaction score.}
\label{fig:fairness}
\vspace{-1.2em}
\end{figure}

To evaluate the effectiveness of the FPS module in promoting fair participant selection, we conduct a controlled experiment on the T-Drive dataset under the \emph{Medium} setting. We fix a pool of 30 workers and consider 60 sensing tasks, where all workers are feasible for task allocation. For each task, the CA selects at most 20 workers, creating a trade-off between sensing performance and participation fairness. We compare four selection strategies: \textit{Uniform Random (UR)} randomly selects workers; \textit{Least-Selected (LS)} prioritizes workers with the fewest historical selections, enforcing strong fairness but ignoring sensing utility; \textit{Coverage-Greedy (CG)} selects workers purely based on marginal coverage gain; and our FPS jointly considers coverage gain and historical participation to balance sensing performance and worker fairness.

Figure~\ref{fig:fairness} compares this four strategies from both fairness and performance perspectives. In particular, we report the cumulative distribution function (CDF) of per-worker selection counts in Figure~\ref{fig:fairness}(b), which shows the distribution of the number of times each worker is selected across 60 tasks. A steeper curve indicates more balanced participation. The results show that CG exhibits severe selection imbalance. A small subset of workers is repeatedly selected while others are rarely chosen, resulting in a long-tailed distribution (Fig.~\ref{fig:fairness}(a, b)). Although CG achieves the highest coverage (Fig.~\ref{fig:fairness}(c)), its lack of diversity leads to the lowest path satisfaction score (Fig.~\ref{fig:fairness}(d)). LS produces the most balanced selection distribution, but this strong fairness constraint leads to the lowest coverage utility. UR shows moderate variability but lacks explicit mechanisms to optimize either fairness or sensing performance. Our FPS method provides a balanced trade-off. The selection distribution is significantly more uniform than CG, while maintaining coverage utility close to CG and higher than UR and LS. In addition, FPS achieves higher path satisfaction than CG, benefiting from more diverse participant assignments. Overall, these results demonstrate that FPS effectively balances sensing performance and participant fairness.

\section{Conclusion}
We presents MAPUS, a human-centric planning framework for participatory urban sensing that incorporates participant profiles and heterogeneous urban context. By modeling participants as autonomous agents and enabling negotiation-based route refinement, the framework moves beyond traditional centralized approaches that optimize sensing utility alone. Experiments on two real-world mobility datasets show that our approach achieves competitive coverage while substantially improving path satisfaction and balancing worker participation. By explicitly integrating human-centric considerations into sensing planning, our framework provides a foundation for more sustainable and socially acceptable urban sensing systems.

\section{Limitations}
Despite the promising results, the proposed framework has several limitations. First, its performance is closely tied to the capability of the underlying LLM, particularly in reasoning about participant preferences and evaluating route satisfaction. Although a classical planner is used to guarantee basic route feasibility, the LLM may still occasionally produce infeasible or low-quality refinements. Second, the participant profiles in the current study are constructed from simulated data rather than real-world user data, which may introduce bias and limit the realism of the resulting behavior patterns. Third, using LLMs to simulate human participants presents a double-edged challenge: while it enables more human-centered and personalized planning, it also raises privacy and ethical concerns, as detailed participant modeling may rely on sensitive personal and behavioral information. Addressing these issues will require future research on stronger feasibility control, real-world profile calibration, and privacy-preserving modeling.

% \section*{Acknowledgement}

% This work is supported by the National Natural Science Foundation of China (No.62402414), the Guangdong Basic and Applied Basic Research Foundation (No. 2025A1515011994), CCF-Kuaishou Large Model Explorer Fund (NO. CCF-KuaiShou 2025018), the Guangzhou Municipal Science and Technology Project (No. 2023A03J0011), the Guangzhou Industrial Information and Intelligent Key Laboratory Project (No. 2024A03J0628), and the Guangdong Provincial Key Lab of Integrated Communication, Sensing and Computation for Ubiquitous Internet of Things (No. 2023B1212010007).
\bibliography{custom}

\clearpage

\appendix

\section{Supplementary Methodological Details}
\label{appendix:methodology_details}

\subsection{Preliminary Definitions}
\label{appendix:problem_setup}

We consider a participatory urban sensing (PUS) program in which a sensing platform aims to collect spatio-temporal data over a predefined urban region and a finite time horizon by leveraging the routine mobility of participating individuals. Following standard formulations in urban sensing \citep{ji2016urban}, each participant reports basic mobility constraints, including origin, destination, speed, and feasible travel time windows. Based on these inputs, the platform jointly decides which participants to recruit and how to assign sensing routes under a global budget constraint, with the objective of maximizing overall sensing utility.

Generally, the sensing region is discretized into a 2D grid with index set $\mathcal{G}=\{(i,j)\mid i=1,\dots,I,\; j=1,\dots,J\}$, and the sensing horizon is divided into time intervals $\mathcal{T}=\{1,\dots,T\}$. Let $\mathcal{U}=\{u_1,\dots,u_N\}$ denote the set of candidate participants. Each participant $u\in\mathcal{U}$ is characterized by the following vector:
\[
u=\langle o_u,\; d_u,\; t_u^{s},\; t_u^{e},\; v_u, \; c_u,\; \boldsymbol{\pi}_u\rangle,
\]
where $o_u,d_u\in\mathcal{G}$ denote the origin and destination, $[t_u^{s},t_u^{e}]$ is the feasible travel time window, $v_u$ is the travel speed, $c_u$ is the recruitment cost, and $\boldsymbol{\pi}_u$ denotes participant-specific preferences. A sensing route for participant $u$ is defined as an ordered sequence
\begin{equation}
r_u=\big((g_1,\tau_1), (g_2,\tau_2), \dots, (g_{L_u},\tau_{L_u})\big),
\label{eq:route_definition_appendix}
\end{equation}
where $g_i \in \mathcal{G}$ and $\tau_i \in \mathcal{T}$, with $\tau_1\le\cdots\le\tau_{L_u}$. The route must satisfy the participant's mobility constraints, including origin--destination consistency and travel-time feasibility.

To model participant selection, we introduce a binary variable
\begin{equation}
x_u=
\begin{cases}
1, & \text{if participant $u$ is selected},\\
0, & \text{otherwise}.
\end{cases}
\label{eq:selection_variable_appendix}
\end{equation}
Given a budget $B$, the total recruitment cost must satisfy
\begin{equation}
\sum_{u\in\mathcal{U}} c_u x_u \le B.
\label{eq:budget_constraint_appendix}
\end{equation}

Combining selection and route assignment, the overall objective is
\begin{equation}
\begin{aligned}
\max_{\{x_u,r_u\}} \quad 
& \phi(\mathbf{A}) + \eta \sum_{u\in\mathcal{U}} x_u\, S(r_u) \\
\text{s.t.} \quad 
& \sum_{u\in\mathcal{U}} c_u x_u \le B
\end{aligned}
\label{eq:overall_objective_appendix}
\end{equation}
where $\phi(\mathbf{A})$ is the sensing utility, $S(r_u)$ is the path satisfaction score, and $\eta$ controls the trade-off between system-level sensing utility and participant-level route satisfaction.

\noindent\textbf{Coverage utility.}
In PUS, the collected data is represented by a tensor $\mathbf{A}\in\mathbb{R}^{I\times J\times T}$, where $\mathbf{A}(i,j,t)$ denotes the amount of data collected at grid cell $(i,j)$ during interval $t$. For a route $r_u$, let $\mathbf{A}_{r_u}$ denote its sensing contribution. Aggregating over selected participants yields $\mathbf{A}=\sum_{u\in\mathcal{U}'} \mathbf{A}_{r_u}$. Following \citet{ji2016urban}, we define
\begin{equation}
\phi(\mathbf{A})=\alpha\,E(\mathbf{A}) + (1-\alpha)\log Q(\mathbf{A}),
\quad \alpha\in[0,1],
\label{eq:coverage_utility_appendix}
\end{equation}
where
\begin{equation}
\setlength{\abovedisplayskip}{6pt} % Adjust space above the equation
\setlength{\belowdisplayskip}{6pt} % Adjust space below the equation
Q(\mathbf{A})=\sum_{i=1}^{I}\sum_{j=1}^{J}\sum_{t=1}^{T} \mathbf{A}(i,j,t)
\label{eq:coverage_count_appendix}
\end{equation}
measures total collected data volume, and $E(\mathbf{A})$ quantifies spatio-temporal balance through entropy across spatial and temporal granularities, as detailed in \citep{ji2016urban}.

\subsection{Route Generation Details}
\label{appendix:route_generation_details}

\begin{algorithm}[!t]
\small
\caption{Preference-Aware Route Generation}
\label{alg:route_planning_appendix}
\begin{algorithmic}[1]
\Require Task setting $\mathcal{T}$; grid map $\mathcal{G}$ with urban attributes; participant schedule $\Sigma=(O,D,t_{\mathrm{dep}},t_{\mathrm{arr}},v)$; participant profile $\mathcal{P}$; max iterations $I_{\max}$; trade-off $\lambda\in[0,1]$
\Ensure Route $\mathcal{R}^*$ (or $\varnothing$ if infeasible)
\State $\mathcal{R}^0 \gets \textsc{TSPTW\_solver}(\Sigma,\mathcal{T})$
\State $u^0 \gets \textsc{Utility}(\mathcal{R}^0; \mathcal{T},\mathcal{P},\mathcal{G},\lambda)$
\State $(\mathcal{R}^*, u^*) \gets (\mathcal{R}^0, u^0)$
\For{$i=1$ to $I_{\max}$}
    \State $\mathcal{R}^i \gets \textsc{LLM\_Refine}(\mathcal{R}^*,\mathcal{T},\Sigma,\mathcal{P},\mathcal{G})$
    \If{\textsc{Valid}$(\mathcal{R}^i,\Sigma,\mathcal{T})$}
        \State $u^i \gets \textsc{Utility}(\mathcal{R}^i;\mathcal{T},\mathcal{P},\mathcal{G},\lambda)$
        \If{$u^i>u^*$}
            \State $(\mathcal{R}^*,u^*) \gets (\mathcal{R}^i,u^i)$
        \EndIf
    \EndIf
\EndFor
\State \Return $\mathcal{R}^*$
\end{algorithmic}
\end{algorithm}

This subsection provides the detailed route-generation procedure used in the first stage of MAPUS. Given a task specification $\mathcal{T}$ and participant schedule $\Sigma=(O,D,t_{\mathrm{dep}},t_{\mathrm{arr}},v)$, where $O/D$ are origin and destination, $t_{\mathrm{dep}}/t_{\mathrm{arr}}$ are earliest departure and latest arrival times, and $v$ is travel speed, each accepted participant first computes a feasible baseline route $\mathcal{R}^0$ and then refines it through LLM-based preference-aware reasoning.
Although the underlying routing problem is NP-hard, the heuristic solver efficiently provides a feasible initialization and narrows the search space for subsequent refinement. Algorithm~\ref{alg:route_planning_appendix} summarizes the procedure. By refining a feasible baseline rather than planning from scratch, the LLM effectively performs a guided local search under hard spatio-temporal constraints. This design combines the feasibility guarantees of classical planning with the flexibility of language-based preference alignment.

\noindent\textbf{Prompt details.}
The participant prompt for preference-aware route generation includes the task description, current route, mobility constraints, participant profile, and grid-level urban attributes, and instructs the LLM to propose route modifications that preserve feasibility while improving route utility. A representative template is in Figure~\ref{fig:participant_prompt}.

\begin{figure}[!t]
\centering
\begin{minipage}{\columnwidth}
\begin{tcolorbox}[
    enhanced,
    colback=gray!10!white,
    colframe=gray!80!white,
    colbacktitle=gray!70!white,
    coltitle=white,
    arc=2mm,
    boxrule=1pt,
    left=1mm,
    right=1mm,
    top=1mm,
    bottom=1mm,
]
\footnotesize

\textbf{\# System Prompt:}\\
You are a worker agent participating in an urban sensing task on a 2D grid of size \textcolor{blue}{\{grid\_size\}}. 
Time steps range from \textcolor{blue}{\{start\_time\}} to \textcolor{blue}{\{end\_time\}}. 
At each time step, you may move up to {speed} grid units in total distance (Manhattan metric), choosing among up, down, left, right, or staying in place. 

\medskip
Your goal is to plan a feasible and realistic route from the origin to the destination, balancing efficiency, personal preferences, and overall convenience.

\medskip
When necessary, you may query grid-level environmental attributes to inform your planning decisions, including land-use characteristics and area safety levels, using the following tools: \\
\quad - \textit{retrieve\_grid\_geographic\_attributes}(x, y) \\
\quad - \textit{retrieve\_grid\_crime\_info}(x, y) \\

\textbf{\# Input Prompt:}\\
Please help optimize the path for the worker traveling from \textcolor{blue}{\{start\_point\}} to \textcolor{blue}{\{end\_point\}} (x, y, t), with a speed limit of \textcolor{blue}{\{speed\}} grids per unit time.

\medskip
Here is an initial path for the worker to begin with:\\
\textcolor{blue}{\{initial\_path\_str\}}

\medskip
The worker has \textcolor{blue}{\{residual\_step\}} residual step(s) remaining after completing the initial path.

\medskip
The worker's profile and preferences are as follows:\\
\textcolor{blue}{\{worker\_profile\}}

\medskip
Here are additional instructions from the coordinator agent regarding path planning:\\
\textbf{Instruction:} \textcolor{blue}{\{instructions\}}\\
Please carefully take the coordinator agent's instructions into account when generating the final path.

\medskip
Please generate an optimized path following these steps: \\
1. Use the residual \textcolor{blue}{\{residual\_step\}} step(s) to add new waypoints along the initial path. These waypoints should prioritize areas that align with the worker's preferences and the attributes of the region. \\
2. Refine the path to better accommodate the worker's preferred areas and time constraints. \\

\textbf{\# Output Format:}
\begin{tcblisting}{
    colback=white,
    colframe=gray!50,
    boxrule=0.5pt,
    arc=1mm,
    listing only,
    left=0mm,right=0mm,top=0mm,bottom=0mm
}
{
  final_path: [(x, y, t), ...],
  explanation: reasoning process...
}
\end{tcblisting}

Do NOT include any comments or non-coordinate text inside \textbf{final\_path}. Do NOT include any text outside the JSON object.
\end{tcolorbox}
\end{minipage}
\caption{Preference-aware route generation prompt.}
\label{fig:participant_prompt}
\vspace{-1em}
\end{figure}

\subsection{Participant Selection Details}
\label{appendix:selection_details}

\begin{algorithm}[!t]
\small
\caption{Fairness-Aware Participant Selection}
\label{alg:worker_selection_appendix}
\begin{algorithmic}[1]
\Require Candidates $\mathcal{U}$ with feasible routes $\{r_u\}_{u\in\mathcal{U}}$; participant profiles $\{\mathcal{P}_u\}_{u\in\mathcal{U}}$; historical assignment table $\{h_u\}_{u\in\mathcal{U}}$; task cost $\{c_u\}_{u\in\mathcal{U}}$; total budget $B$; trade-off $\beta\in[0,1]$; tie threshold $\epsilon$
\Ensure Selected participants $\mathcal{S}$
\State $\mathcal{S}\gets\varnothing$, $B_{\mathrm{rem}}\gets B$
\While{$B_{\mathrm{rem}} > 0$}
    \State $\mathcal{U}_{\mathrm{feas}} \gets \{u\in\mathcal{U}\setminus\mathcal{S}\mid c_u \le B_{\mathrm{rem}}\}$
    \If{$\mathcal{U}_{\mathrm{feas}}=\varnothing$} \textbf{break} \EndIf
    \ForAll{$u\in \mathcal{U}_{\mathrm{feas}}$}
        \State $\Delta \phi(u) = \phi\!\left(\bigcup_{w\in \mathcal{S}\cup\{u\}} r_w\right) - \phi\!\left(\bigcup_{w\in \mathcal{S}} r_w\right)$
        \State $F(u)=\frac{1}{1+h_u}$
    \EndFor
    \State Normalize $\{\Delta \phi(u)\}$ and $\{F(u)\}$ to obtain $\widetilde{\Delta \phi}(u)$ and $\widetilde{F}(u)$
    \State $J(u)\gets \beta\cdot \widetilde{\Delta \phi}(u) + (1-\beta)\cdot \widetilde{F}(u)$ for all $u\in \mathcal{U}_{\mathrm{feas}}$
    \State $\mathcal{T}\gets \{u\in \mathcal{U}_{\mathrm{feas}} \mid J(u)\ge \max_{w\in \mathcal{U}_{\mathrm{feas}}} J(w)-\epsilon\}$
    \If{$|\mathcal{T}|=1$}
        \State $u^*\gets$ the unique element in $\mathcal{T}$
    \Else
        \State $u^*\gets \textsc{LLM\_TieBreak}(\mathcal{T}, \{\mathcal{P}_u\}_{u\in\mathcal{T}})$
    \EndIf
    \State $\mathcal{S}\gets \mathcal{S}\cup\{u^*\}$
    \State $B_{\mathrm{rem}}\gets B_{\mathrm{rem}}-c_{u^*}$
\EndWhile
\State \Return $\mathcal{S}$
\end{algorithmic}
\end{algorithm}

Algorithm~\ref{alg:worker_selection_appendix} summarizes the selection procedure. Starting from an empty set, the coordinator iteratively selects participants under the remaining budget constraint. At each step, all budget-feasible candidates are evaluated by their marginal coverage gain and long-term fairness score, which are normalized and combined into a composite selection score. The candidate with the highest score is selected, while ties are resolved through an LLM-based tie-breaking mechanism that considers participant profiles. The process terminates when the budget is exhausted or no feasible participant.

\noindent\textbf{Tie-breaking prompt.}
To resolve ties in the selection score, the coordinator invokes an LLM-based tie-breaking mechanism. When multiple candidates obtain indistinguishable scores, the LLM is prompted to reason over their profiles and historical participation records to select one participant based on high-level fairness considerations that are not explicitly modeled in the numerical objective. Figure~\ref{fig:selection_prompt} shows the prompt template used in this step. The prompt provides the set of tied candidates and asks the LLM to choose one participant according to fairness and equity principles.

\begin{figure}[!t]
\centering
\begin{minipage}{\columnwidth}

\begin{tcolorbox}[
    enhanced,
    colback=gray!10!white,
    colframe=gray!80!white,
    arc=2mm,
    boxrule=1pt,
    left=1mm,
    right=1mm,
    top=1mm,
    bottom=1mm,
]

\footnotesize

\textbf{\# System Prompt:}\\
You are a fairness-aware assistant for participant selection in an urban sensing task.

\medskip

You are given a set of candidate participants who are \textbf{tied} according to quantitative selection criteria and all satisfy budget and feasibility constraints. For each candidate, you are provided with a short profile containing background and historical participation information.

\medskip

Your task is to select one participant by considering high-level fairness and equity principles that are not captured by numerical scores, such as prior participation opportunities or socioeconomic context.

\medskip

\textbf{\# Input Prompt:}\\
The following candidate participants are tied under the current selection criteria. All candidates satisfy feasibility and budget constraints. For each candidate, a profile is provided including background information and historical participation records. Select one participant from the list based on fairness considerations.

\quad - \textbf{Candidates: } 
\textcolor{blue}{$\{(u_1, P_{u_1}), \dots, (u_n, P_{u_n})\}$}

\medskip

\textbf{\# Output:} Return the ids of the selected participants.

\end{tcolorbox}

\end{minipage}
\caption{Prompt template used for LLM-based tie-breaking in fairness-aware participant selection..}
\label{fig:selection_prompt}
\vspace{-1em}
\end{figure}

\subsection{Negotiation and Refinement Details}
\label{appendix:negotiation_details}

\begin{algorithm}[!t]
\small
\caption{Negotiation-Based Route Refinement}
\label{alg:negotiation_refinement_appendix}
\begin{algorithmic}[1]
\Require Task setting $\mathcal{T}$; selected participants $\mathcal{S}$ with routes $\{r_u\}_{u\in\mathcal{S}}$; participant schedules $\{\Sigma_u\}$; participant profiles $\{\mathcal{P}_u\}$; semantic grid map $\mathcal{G}$; overlap threshold $\tau$; max negotiation rounds $K$; max pair attempts $N$
\Ensure Refined routes $\{r_u\}_{u\in\mathcal{S}}$ and failed-pair set $\mathcal{F}$
\State $\mathcal{F}\gets\varnothing$
\For{$n=1$ to $N$}
   \State $(u,v)\gets \arg\max_{(i,j):\,i\neq j,\,(i,j)\notin \mathcal{F}} \mathrm{Overlap}(r_i,r_j)$
    \State $o_{\max}\gets \mathrm{Overlap}(r_u,r_v)$
    \If{$o_{\max}<\tau$} \textbf{break} \EndIf
    \State $f_u\gets \varnothing,\ f_v\gets \varnothing$
    \State $I_u\gets \varnothing,\ I_v\gets \varnothing$
    \State $\textsc{Success}\gets \textbf{false}$
    \For{$k=1$ to $K$}
        \State $(\hat{r}_u,\hat{r}_v,I_u,I_v) \gets$
        \Statex \quad $\textsc{CA\_Refine}(\mathcal{T},\mathcal{G},r_u,r_v,P_u,P_v,f_u,f_v)$
        
        \State $(a_u,f_u) \gets$
        \Statex \quad $\textsc{PA\_Feedback}(\mathcal{T},\mathcal{G},u,\hat{r}_u,\Sigma_u,P_u,I_u,f_v)$
        
        \State $(a_v,f_v) \gets$
        \Statex \quad $\textsc{PA\_Feedback}(\mathcal{T},\mathcal{G},v,\hat{r}_v,\Sigma_v,P_v,I_v,f_u)$
        \If{$a_u$ \textbf{and} $a_v$}
            \State $(r_u,r_v)\gets(\hat{r}_u,\hat{r}_v)$
            \State $\textsc{Success}\gets \textbf{true}$
            \State \textbf{break}
        \EndIf
    \EndFor
    \If{\textbf{not} $\textsc{Success}$}
        \State $\mathcal{F}\gets \mathcal{F}\cup\{(u,v)\}$
    \EndIf
\EndFor
\State \Return $\{r_u\}_{u\in\mathcal{S}},\ \mathcal{F}$
\end{algorithmic}
\end{algorithm}

Algorithm~\ref{alg:negotiation_refinement_appendix} summarizes the negotiation-based route refinement procedure. Starting from the routes generated by the selected participants, the coordinator iteratively identifies the pair of routes with the highest overlap that has not previously failed negotiation. For each selected pair, the coordinator proposes refined routes, while participant agents evaluate the proposals and return accept/reject decisions together with feedback. If both participants accept the proposal, the refined routes are adopted; otherwise, the coordinator updates the proposal for a limited number of negotiation rounds. Pairs that fail to reach agreement are recorded and skipped in subsequent iterations. The procedure terminates when the overlap falls below a predefined threshold or the maximum number of pair attempts is reached.

\noindent\textbf{Prompt for the coordinator agent in the negotiation process.} Fig.~\ref{fig:coordinator_prompt} presents the prompt used for the coordinator agent in the negotiation stage. The coordinator serves as a mediator that proposes route refinements for participants with highly overlapping sensing paths. Given the task setting, participant routes, profiles, and feedback from previous rounds, the prompt guides the agent to generate feasible route adjustments that reduce spatial redundancy while respecting participant preferences and maintaining negotiation-based acceptance.

\begin{figure}[!t]
\centering
\small
\begin{minipage}{\columnwidth}

\begin{tcolorbox}[
    enhanced,
    colback=gray!10!white,
    colframe=gray!80!white,
    arc=2mm,
    boxrule=1pt,
    left=1mm,
    right=1mm,
    top=1mm,
    bottom=1mm,
]

\footnotesize

\textbf{\# System Prompt:}\\
You are a coordinator agent responsible for refining sensing routes through negotiation in an urban sensing task. You act as a mediator between participant agents and cannot enforce route changes; all refinements must be accepted by the participants.

\medskip
Your objective is to propose refined routes that: \\
- reduce spatial overlap between participants; \\
- remain feasible under task setting and constraints; \\
- respect participant preferences as much as possible.

\medskip
You should iteratively improve your proposals based on feedback from previous negotiation rounds.

\medskip
\textbf{Task setting:} \textcolor{blue}{\{task\_setting\}}

\medskip
When needed, you may query grid-level attributes to assess route feasibility and safety with following tools: \\
\quad - \textit{retrieve\_grid\_geographic\_attributes}(x, y) \\
\quad - \textit{retrieve\_grid\_crime\_info}(x, y) \\
Use the retrieved information to guide route adjustments.

\medskip
\textbf{\# Input Prompt:}\\
Two participants have been identified with highly overlapping routes.

\medskip
\textbf{Participant \(u\):} \textcolor{blue}{\{ID: \{u\}, Route: \{route\_u\}, Profile: \{profile\_u\}, Feedback: \{feedback\_u\}\}}

\medskip
\textbf{Participant \(v\):} \textcolor{blue}{\{ID: \{v\}, Route: \{route\_v\}, Profile: \{profile\_v\}, Feedback: \{feedback\_v\}\}}

\medskip
Your task is to propose refined routes for both participants that reduce overlap while keeping each participant’s origin and destination unchanged. Adjustments should focus on overlapping or middle segments and preserve route feasibility and realism.

\medskip
You may optionally provide short incentive messages explaining the rationale behind the proposed refinements to each participant.

\medskip
\textbf{\# Output Format:}
\begin{tcblisting}{
    colback=white,
    colframe=gray!50,
    boxrule=0.5pt,
    arc=1mm,
    listing only,
    left=0mm,
    right=0mm,
    top=0mm,
    bottom=0mm
}
{
  "refined_routes": {
    u: [(x, y, t), ...],
    v: [(x, y, t), ...]
  },
  "incentives": {
    u: incentive to accept refined route...,
    v: incentiveto accept refined route...
  }
}
\end{tcblisting}

Do NOT include any text outside the JSON object.

\end{tcolorbox}

\end{minipage}
\caption{Prompt for the coordinator agent in the negotiation process.}
\label{fig:coordinator_prompt}
\end{figure}

\noindent\textbf{Prompt for the participant agent in the negotiation process.} Fig.~\ref{fig:participant_prompt} shows the prompt used for participant agents in the negotiation stage. Based on the task setting, personal schedule, and individual preferences, the prompt instructs each participant agent to evaluate the feasibility and acceptability of route refinements proposed by the coordinator. The agent returns a binary agreement decision together with brief feedback explaining the rationale, which enables iterative negotiation and refinement of sensing routes.

\begin{figure}[!t]
\small
\centering
\begin{minipage}{\columnwidth}

\begin{tcolorbox}[
    enhanced,
    colback=gray!10!white,
    colframe=gray!80!white,
    arc=2mm,
    boxrule=1pt,
    left=1mm,
    right=1mm,
    top=1mm,
    bottom=1mm,
]

\footnotesize

\textbf{\# System Prompt:}\\
You are a participant agent in an urban sensing task. Your role is to evaluate route refinements proposed by the coordinator agent and decide whether to accept them.

\medskip
Your decision should consider the following information: \\
- \textbf{Task setting:} \textcolor{blue}{\{task\_setting\}} \\
- \textbf{Your schedule:} \textcolor{blue}{\{schedule\}} \\
- \textbf{Your profile and preferences:} \textcolor{blue}{\{participant\_profile\}}

\medskip
When making decision, consider the following criteria: \\
- feasibility considering your schedule and task constraints; \\
- alignment with your personal preferences; and \\
- the reasonableness of the coordinator's explanation.

\medskip
You may tolerate minor inconvenience if the proposal is well justified. If you reject a proposal, clearly state the primary reason for rejection.

\medskip
\textbf{\# Input Prompt:}\\
A refined route has been proposed for you. \\
- \textbf{Proposed route:} \textcolor{blue}{\{proposed\_route\}} \\
- \textbf{Original route:} \textcolor{blue}{\{original\_route\}} \\
- \textbf{Coordinator message (if any):} \textcolor{blue}{\{incentive\_message\}} \\
- \textbf{Feedback from previous negotiation rounds (if any):} \textcolor{blue}{\{feedback\_memory\}}

\medskip
Your task is to determine whether to accept the proposed route.

\medskip
\textbf{\# Output Format:}

\begin{tcblisting}{
    colback=white,
    colframe=gray!50,
    boxrule=0.5pt,
    arc=1mm,
    listing only,
    left=0mm,
    right=0mm,
    top=0mm,
    bottom=0mm
}
{
  agreement: True/False,
  feedback: Brief explanation of your decision. If rejected, clearly state the main reason.
}
\end{tcblisting}

Do NOT include any text outside the JSON object.

\end{tcolorbox}

\end{minipage}
\caption{Prompt for the participant agent in the negotiation process.}
\label{fig:participant_prompt}
\end{figure}

\clearpage 

\section{Supplementary Experimental Details}
\label{appendix:experiment_details}

\subsection{Urban Attributes}
\label{appendix:urban_attributes}

To characterize region-level contextual signals, we incorporate two grid-level attributes: land-use distribution and crime statistics. Land-use composition is estimated by aggregating points of interest (POIs) within each grid cell. POIs are grouped into six land-use categories, and their relative frequencies are used to approximate the functional composition of each grid cell.

For the T-Drive task region, we additionally incorporate crime statistics to capture spatial heterogeneity in sensing demand and operational risk. Crime data are obtained from~\citep{zhang2025llm} and aggregated over the same spatial grid. Grid cells with fewer than 10 recorded incidents are excluded to reduce noise. Since reliable crime data are unavailable for the Grab-Posisi task region, grid attributes for that dataset consist solely of POI-based land-use distributions.

\begin{figure}[!t]
    \centering
    \includegraphics[trim={0cm 8.5cm 4cm 0cm}, clip, width=\linewidth]{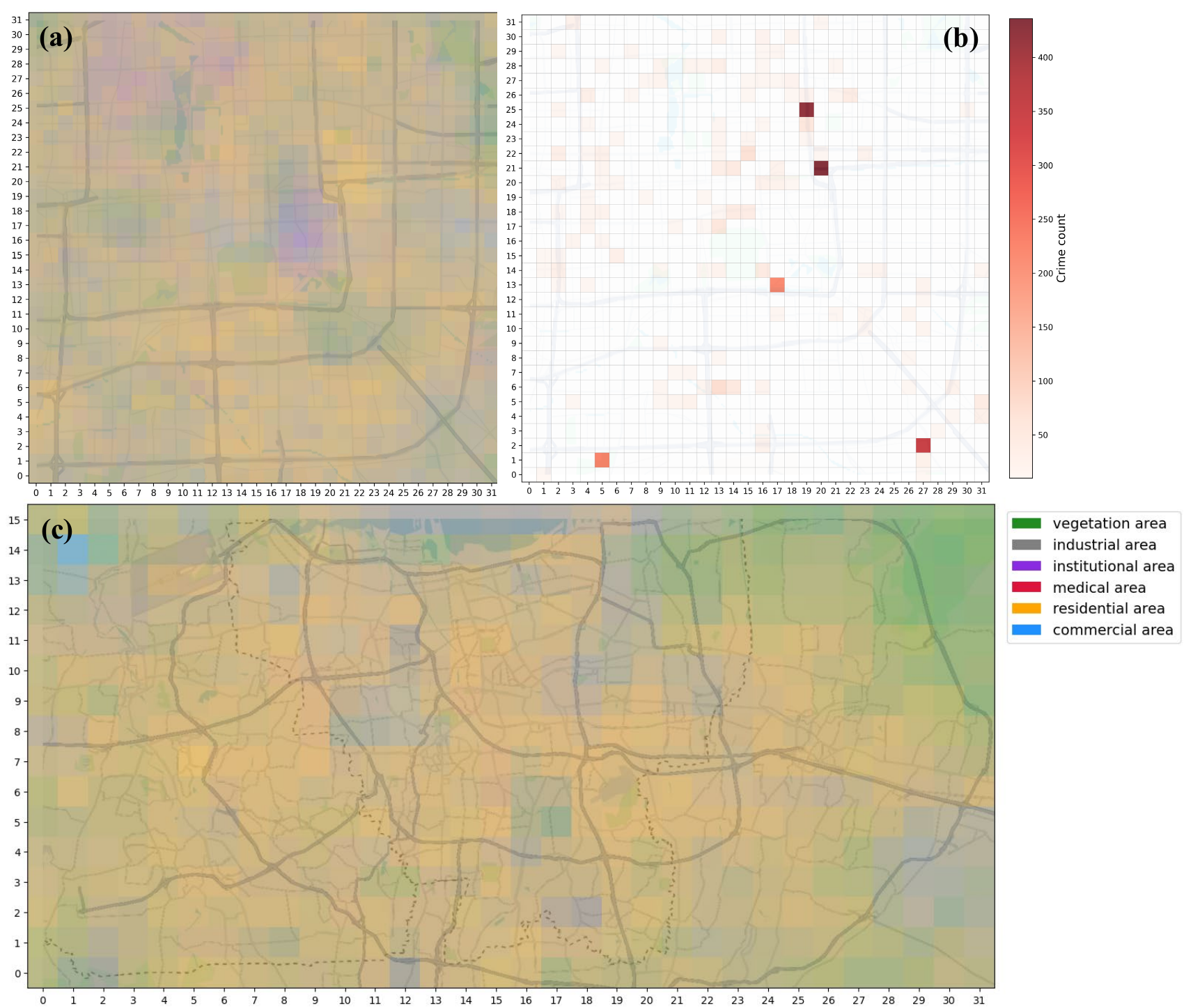}
    \caption{Visualization of land-use and crime distribution in task regions. 
    (a) Land-use distribution in the T-Drive task area derived from aggregated POI categories. 
    (b) Spatial distribution of crime counts in the T-Drive task area aggregated over grid cells. 
    (c) Land-use distribution in the Grab-Posisi task area.}
\label{fig:data}
\vspace{-1.0em}
\end{figure}

\subsection{Participant Profiles Construction}
\label{appendix:participant_profiles}

\begin{figure}[!t]
\small
\centering
\begin{minipage}{\columnwidth}

\begin{tcolorbox}[
    enhanced,
    colback=gray!10!white,
    colframe=gray!80!white,
    arc=2mm,
    boxrule=1pt,
    left=1mm,
    right=1mm,
    top=1mm,
    bottom=1mm
]

\footnotesize

\textbf{\# Worker Profile Template}

\begin{verbatim}
{
  worker_id: ID,
  gender: {male | female | non-binary},
  age: integer,
  age_group: {young | middle_aged | senior},
  economic_status: {poor | low | middle 
                    | high | wealthy},
  hobbies: [hobby_1, hobby_2, ...],
  type: {eco_enthusiast | city_dweller | 
         industrial_worker | community_helper 
         | explorer},
  description: summary of preferences,
  preferences: {
      vegetation_area: float,
      industrial_area: float,
      institutional_area: float,
      medical_area: float,
      residential_area: float,
      commercial_area: float
  },
  speed: float
}
\end{verbatim}

\end{tcolorbox}

\end{minipage}
\caption{Template used to construct synthetic participant profiles.}
\label{fig:worker_profile_template}
\vspace{-1.0em}
\end{figure}

To model heterogeneous participant behavior in the urban sensing task, we construct lightweight synthetic participant profiles that combine demographic attributes, behavioral traits, and geographically grounded preferences. Each participant is instantiated with a structured profile (as shown in Figure~\ref{fig:worker_profile_template}) injected into the worker agent prompt to guide preference-aware route planning.

Demographic attributes (e.g., gender, age, and economic status) are sampled from predefined distributions to approximate realistic population diversity. Behavioral traits are represented by a random subset of hobbies that capture individual lifestyle differences. Geographic preferences are modeled through a land-use affinity vector over urban land-use categories. The \texttt{type} field assigns each participant to a preference archetype associated with a land-use preference vector and demographic priors (age group and economic status). Archetypes are sampled probabilistically according to the compatibility between participant demographics and archetype priors.

We define five representative archetypes: \ding{172} \textit{Eco-Enthusiast}: Prefers green spaces and quiet neighborhoods; \ding{173} \textit{City Dweller}: Favors dense residential and commercial areas; \ding{174} \textit{Industrial Worker}: Frequently operates near industrial zones; \ding{175} \textit{Community Helper}: Prefers regions near hospitals, schools, and community facilities; \ding{176} \textit{Explorer}: Exhibits balanced preferences across different regions. Each archetype corresponds to a preference vector over land-use categories (vegetation, industrial, institutional, medical, residential, and commercial). During route generation, this profile biases the worker agent toward regions that better match individual preferences while still satisfying mobility and coordination constraints.

\begin{figure*}[!ht]
    \centering
    \includegraphics[width=\linewidth]{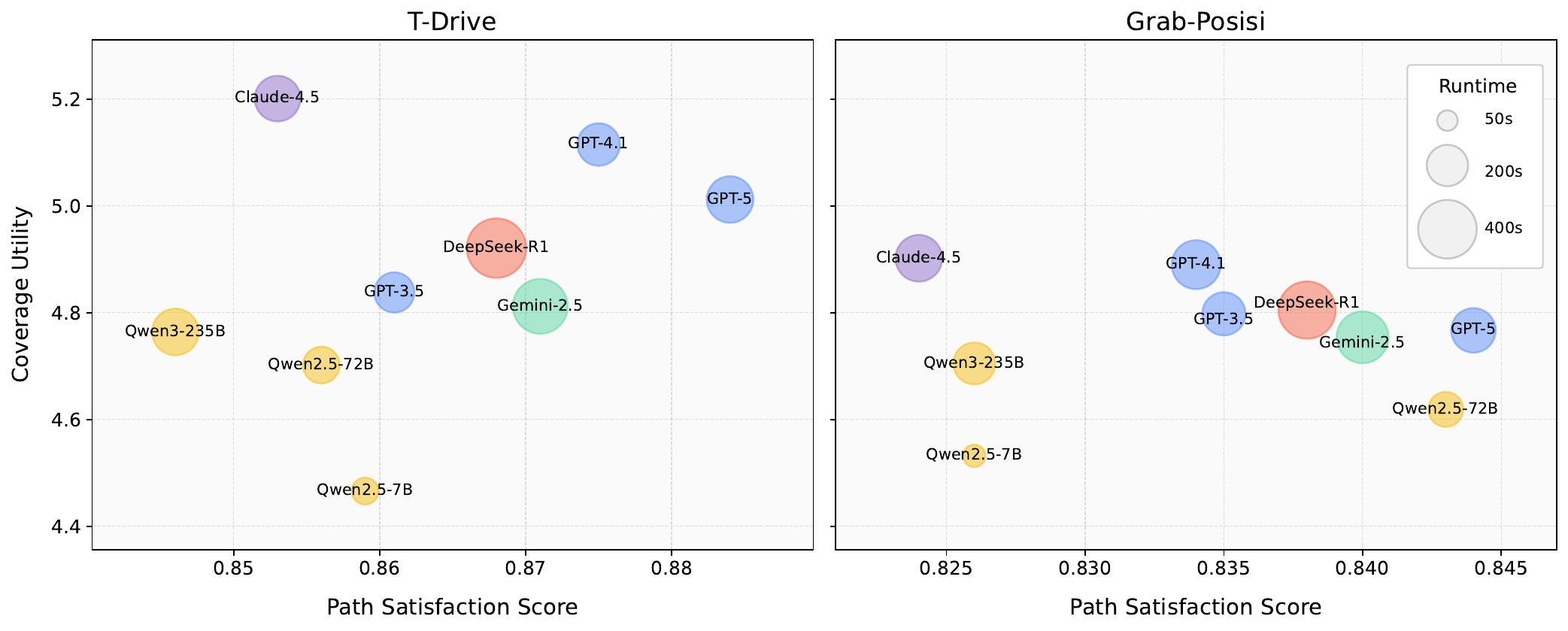}
     \caption{Performance trade-offs of different base LLMs under the \emph{Medium} setting (20 tasks). The x-axis shows path satisfaction score (PSS) and the y-axis shows coverage utility. Bubble size indicates runtime. Results are shown for T-Drive (left) and Grab-Posisi (right).}
\label{fig:llm_bubble_tradeoff}
\end{figure*}

\subsection{Baselines}
\label{appendix:baselines}

We compare MAPUS with six representative planners. Five methods (RN, TVPG, TCPG, MSA, and MSAGI) follow the implementations in~\citep{wang2024urban}, while \emph{GraphDP} is based on~\citep{ji2016urban}. These baselines cover diverse strategies including random assignment, greedy heuristics, metaheuristic search, and dynamic programming.

\begin{itemize}[leftmargin=*,itemsep=0.1em,topsep=0em]

\item \textbf{Random (RN).}
A stochastic baseline where each worker follows a feasible route generated using the \textit{Nearest Neighbor} heuristic. Sensing tasks are randomly inserted into feasible positions until the worker’s budget is exhausted.

\item \textbf{Task Value Priority Greedy (TVPG).}
Tasks are greedily inserted into worker routes to maximize marginal coverage gain. When multiple tasks yield equal gains, the task with lower incentive cost is selected.

\item \textbf{Task Cost Priority Greedy (TCPG).}
Similar to TVPG but prioritizes tasks with lower incentive cost. Coverage gain is used as a secondary criterion to break ties.

\item \textbf{Multi-Start Simulated Annealing (MSA).}
A metaheuristic method that performs local search operations (e.g., task insertion, swapping, and route reversal) under a simulated annealing acceptance rule. Multiple random restarts improve robustness.

\item \textbf{Multi-Start Simulated Annealing with Greedy Initialization (MSAGI).}
An enhanced version of MSA initialized with TVPG, which improves convergence and solution quality.

\item \textbf{Graph-based Dynamic Programming (GraphDP).}
Models each worker’s sensing process as a time-expanded location graph and applies dynamic programming to maximize spatial–temporal coverage. A worker replacement procedure further improves overall coverage.

\end{itemize}

\subsection{Runtime Analysis on Main Results.}

\begin{table*}[!t]
\centering
\small
\renewcommand{\arraystretch}{1.1}
\begin{tabular}{c|ccc|ccc}
\Xhline{1.2pt}
\multirow{2}{*}{Method} 
& \multicolumn{3}{c|}{Avg. runtime on \textbf{T-Drive} (s)}
& \multicolumn{3}{c}{Avg. runtime on \textbf{Grab-Posisi} (s)} \\
\cline{2-7}
& Small & Medium & Large
& Small & Medium & Large \\
\Xhline{1.2pt}

RN     
& \textbf{6.754e-4} & \textbf{4.453e-4} & \textbf{1.142e-3}
& \textbf{2.999e-4} & \textbf{9.423e-4} & \textbf{1.568e-3} \\

TVPG   
& 0.010 & 0.048 & 0.472
& 0.0039 & 0.0212 & 0.266 \\

TCPG   
& 0.012 & 0.112 & 1.155
& 0.0044 & 0.0394 & 0.388 \\

MSA    
& 0.040 & 0.077 & 0.217
& 0.0319 & 0.0616 & 0.157 \\

MSAGI  
& 0.089 & 0.322 & 2.442
& 0.0509 & 0.168 & 1.502 \\

GraphDP
& 0.160 & 7.763 & 147.94
& 0.0366 & 0.798 & 15.22 \\

GPT-4.1-mini
& 158.72 & 187.15 & 210.32
& 125.52 & 156.38 & 183.91 \\

\textbf{MAPUS (Ours)}
& 232.12 & 254.66 & 283.41
& 228.62 & 243.06 & 262.74 \\

\Xhline{1.2pt}
\end{tabular}
\caption{Runtime comparison across datasets and task scales. The main values denote the average runtime (seconds), and the lower-right numbers indicate the variance across 20 task instances.}
\label{tab:runtime_comparison}
\end{table*}

To complement the main results in Tables~\ref{tab:tdrive_overall} and~\ref{tab:grab_overall}, we further analyze the runtime of different methods; the results are reported in Table~\ref{tab:runtime_comparison}. Overall, the compared methods exhibit markedly different runtime characteristics. Random and greedy baselines incur negligible computation cost, while metaheuristic approaches such as MSA and MSAGI show steadily increasing runtime as the task scale grows. GraphDP, which relies on dynamic programming, also suffers from a clear scalability issue: on T-Drive, its runtime rises from sub-second levels in small instances to over 140 seconds in the \emph{Large} setting, reflecting the rapid growth of the underlying state space. Both MAPUS and the single-LLM  baseline incur substantially higher runtime than classical baselines due to the use of LLM inference. Among them, MAPUS is generally slower because it involves multi-agent cooperation.

Importantly, the reported runtime of MAPUS and single-LLM  baseline measure the end-to-end execution of the complete workflow, from API invocation to final route assignment. As such, it reflects not only raw computation time but also practical system factors such as network latency and service-side load balancing. Although these measurements do not directly represent pure compute consumption, they better capture the latency experienced in realistic deployment scenarios where LLMs are accessed through remote APIs.

Despite the higher cost, the runtime of MAPUS remains relatively stable across task scales, suggesting that its overhead is dominated by a bounded number of LLM-based reasoning and coordination steps rather than combinatorial growth of the solution space. Compared with the single-LLM baseline, MAPUS trades additional runtime for a more structured planning process, yielding better solution quality and stability. Since PUS planning is typically performed offline, this added cost is acceptable in practice, and the multi-agent design also offers opportunities for distributed execution in future deployments.

\subsection{Results on Different Base LLMs}

To examine the impact of the base LLM on solution quality and runtime, we evaluate several models within our framework under the \emph{Medium} setting (20 tasks) on both T-Drive and Grab-Posisi. Figure~\ref{fig:llm_bubble_tradeoff} visualizes the trade-off between coverage utility, path satisfaction score (PSS), and runtime.

\paragraph{Performance Scaling and Robustness.}
Across both datasets, a consistent scaling trend emerges: higher-capacity LLMs generally gravitate toward the upper-right quadrant, simultaneously maximizing coverage utility and path satisfaction. However, we observe a distinct difference in sensitivity between these two metrics. While coverage utility exhibits a wide spread across different backbones, path satisfaction scores remain relatively clustered. This suggests that while global sensing optimization is highly dependent on model capacity, our framework's ability to maintain preference alignment is remarkably robust, even when instantiated with lightweight models like Qwen2.5-7B.

\paragraph{The Efficiency–Performance Trade-off.}
The bubble sizes in Figure~\ref{fig:llm_bubble_tradeoff} highlight a stark efficiency trade-off. Frontier models such as DeepSeek-R1 and Gemini-2.5-Pro deliver peak objective performance but incur substantial end-to-end latency. In contrast, lightweight open-source models offer a compelling alternative for latency-sensitive deployments; they achieve competitive satisfaction scores with significantly lower runtime, albeit at the cost of some global coverage. This bifurcation allows the framework to be tailored to specific operational constraints, whether the priority is exhaustive optimization or real-time response.

\paragraph{Model-Family Inductive Biases.}
Beyond raw capacity, we identify systematic variations in how different model families approach the multi-objective problem. Claude models consistently lean toward higher coverage utility, suggesting a stronger inductive bias for global combinatorial optimization. Conversely, the GPT series tends to prioritize path satisfaction, reflecting a potential emphasis on fine-grained instruction following and individual preference alignment. These discrepancies likely stem from the divergent training data and RLHF strategies employed by different providers.

Overall, while the framework's performance is inevitably influenced by the choice of LLM, it demonstrates high adaptability across a broad spectrum of models. This flexibility enables practitioners to strategically select a backbone that balances solution quality, computational budget, and specific objective priorities.

\subsection{Case Studies for Dynamic Disturbances}

\begin{figure*}
    \centering
    \includegraphics[trim={1cm 0.5cm 1cm 0.5cm}, clip, width=\linewidth]{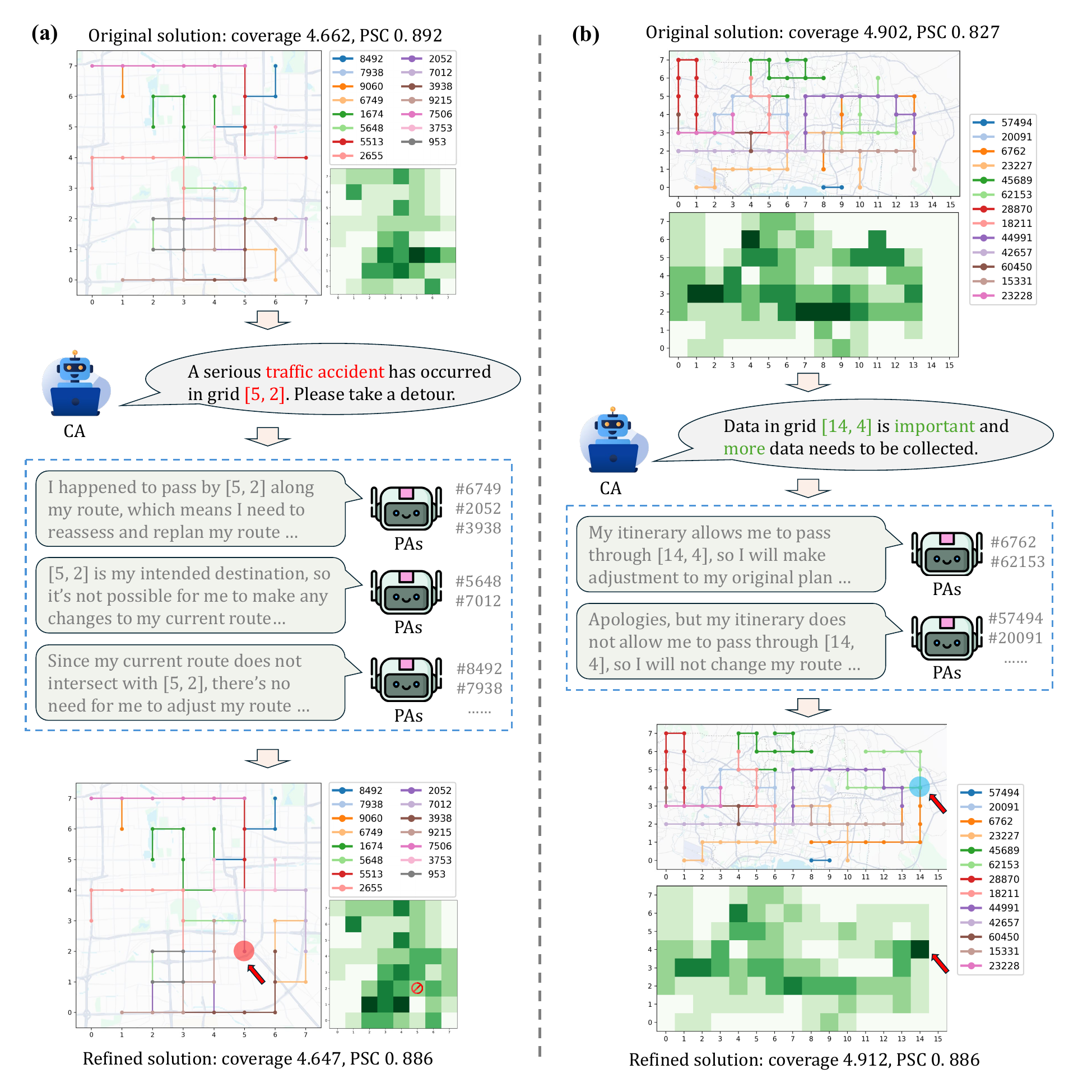}
     \caption{
     Case studies of route adaptation under dynamic disturbances. 
(a) A traffic accident causing grid [5, 2] blockage in the T-Drive dataset. 
(b) A high-priority data collection task in grid [14, 4] from the Grab-Posisi dataset. 
In each case, the coordinator agent (CA) broadcasts dynamic information to participant agents (PAs). PAs evaluate whether their routes are affected and adjust accordingly if necessary. We report the coverage utility and path satisfaction score for the initial and refined solutions, with the areas modified in response to the dynamic disturbances highlighted by red arrows.
}
\label{fig:case_study}
\end{figure*}

MAPUS also possesses the ability to adapt to dynamic disturbances, which is challenging for traditional methods. To demonstrate this, we introduce the coordinator agent (CA) that broadcasts real-time event information and allows participant agents (PAs) to adjust their previously planned routes, thereby evolving the initial solution. In these experiments, only the preference-aware route generation module and the negotiation-based route refinement module are activated, which means that the set of selected workers remains fixed, and participants adjust only their planned routes in response to dynamic events. We present two representative case studies in Figure~\ref{fig:case_study} to illustrate this adaptation. Case (a) considers a traffic accident that blocks a specific grid region, while case (b) introduces a high-priority data collection task.

Across both cases, we observe that different participant agents adopt distinct adaptation strategies in response to the disturbances. Upon receiving the broadcasted information, each PA first evaluates whether the disturbance truly affects its planned route. If so, the PA adjusts its route to accommodate the new constraints, as illustrated by participant agents with IDs \#6749, \#2052, and \#3938 in case (a), and \#6762 and \#62153 in case (b). In contrast, when a PA determines that the disturbance does not affect its planned route, it tends to preserve the original plan to avoid unnecessary deviations. Moreover, we observe some edge cases where a disturbance affects a PA’s route but cannot be reasonably avoided. For example, in case (a), participant agents \#5648 and \#7012 have destinations located within the grid region affected by the accident. In such situations, the agents correctly decide to maintain their original routes rather than making infeasible adjustments. These behaviors highlight the strong reasoning and language understanding capabilities of the LLM-driven agents, enabling them to make context-aware decisions in a human-like manner.

We also observe changes in both the coverage utility and the path satisfaction score between the initial and refined solutions. These variations are expected, as the introduction of dynamic disturbances imposes additional constraints on the system and alters the feasible routing space. Notably, in case (b), the refined solution achieves a higher coverage utility than the initial one. This suggests that the original solution was only locally optimal, and the newly introduced disturbance helped break the previous equilibrium, allowing the system to escape suboptimal configurations and explore more advantageous coordination patterns. In this sense, dynamic events can act as a catalyst for re-optimization rather than merely a disruption. Overall, these case studies underscore the adaptability of our framework to varying environmental conditions, demonstrating that it can maintain effective and coherent decision-making even with unexpected disturbances while continuing to balance sensing performance and participant satisfaction.

\end{document}